\documentclass[10pt,twocolumn,letterpaper]{article}

\usepackage[pagenumbers]{cvpr}      
\usepackage{graphicx}
\usepackage{amsmath}
\usepackage{amssymb}
\usepackage{bbm}
\usepackage{tabularx}
\usepackage{booktabs}
\usepackage[normalem]{ulem}
\usepackage{caption}
\usepackage{mathrsfs}
\usepackage{mathtools}
\usepackage{rotating}
\usepackage{multirow}
\usepackage{lipsum}
\usepackage[dvipsnames]{xcolor}
\usepackage{enumitem}
\usepackage{inconsolata}
\usepackage{pifont}
\usepackage{soul}
\usepackage{siunitx}
\usepackage{subcaption}
\usepackage{float}
\usepackage[accsupp]{axessibility}  
\usepackage{xcolor}
\usepackage{balance}
\usepackage{makecell}
\usepackage{colortbl} 
\usepackage{textcomp}
\usepackage[normalem]{ulem}       
\usepackage{contour}

\contourlength{0.8pt}

\newcommand{\myuline}[1]{%
  \uline{\phantom{#1}}
  \llap{\contour{white}{#1}}
}
\usepackage{algorithm}
\usepackage{algpseudocode}
\usepackage{xspace}

\newcommand{\TODO}[1]{\textbf{\color{red}[TODO: #1]}}
\useunder{\uline}{\ul}{}

\newcommand{\modelname}{DUET-VLM}

\renewcommand{\paragraph}[1]{\vspace{1mm}\noindent\textbf{#1}}

\newcommand{\AMDGPU}{AMD~Instinct\textsuperscript{\texttrademark}}

\newcommand{\bX}{\mathbf{X}}

\newcommand{\bx}{\mathbf{x}}
\newcommand{\bz}{\mathbf{z}}
\newcommand{\bbR}{\mathbb{R}}
\newcommand{\avv}{A_{v2v}}
\newcommand{\atv}{A_{t2v}}

\newcommand{\Xr}{\bX_{\text{res}}}
\newcommand{\Xc}{\bX_{\text{comp}}}
\newcommand{\Xd}{\bX_{\text{dom}}}
\newcommand{\Xo}{\bX_{\text{out}}}
\newcommand{\topk}{\text{TopK}}
\newcommand{\call}{(\textrm{C+all})\xspace}
\newcommand{\cs}{(\textrm{C+S})\xspace}
\newcommand{\crm}{(\textrm{C})\xspace}

\newcommand{\MD}{\mathcal{D}}

\newcommand{\MN}{\mathcal{N}}

\newcommand{\MV}{\mathcal{V}}
\newcommand{\MVl}{\mathcal{V}^{(l)}}
\newcommand{\MS}{\mathcal{S}}
\newcommand{\MT}{\mathcal{T}}

\newcommand{\MC}{\mathcal{C}}
\newcommand{\MR}{\mathcal{R}}

\definecolor{lightpink}{HTML}{FFD6BA}
\definecolor{lightyellow}{HTML}{FFF9BD} 
\definecolor{ly}{HTML}{FAD9D5} 
\definecolor{lg}{HTML}{DAE8FC} 

\definecolor{darkred}{rgb}{0.5, 0.0, 0.13}
\definecolor{royalblue}{HTML}{002855}
\definecolor{yorange}{HTML}{FFA445}
\definecolor{g1}{HTML}{2A7208}
\definecolor{g2}{HTML}{4A9C13}
\definecolor{g3}{HTML}{82C720}
\definecolor{purpletext}{HTML}{8856A7}

\newcommand{\confName}{[Conference Name]}

\usepackage{placeins}
\usepackage{microtype}
\usepackage{suffix} 
\WithSuffix\def\R1{{\color{red}{kUJb}}}
\WithSuffix\def\R2{{\color{blue}{cSX8}}}
\WithSuffix\def\R3{{\color{OliveGreen}{s2kr}}}
\WithSuffix\def\R4{{\color{RawSienna}{B1CR}}}
\WithSuffix\def\R5{{\color{orange}{pYd1}}}
\WithSuffix\def\R6{{\color{violet}{oW6G}}}
\WithSuffix\def\R7{{\color{teal}{1tRd}}}

\renewcommand{\TODO}[1]{}

\definecolor{cvprblue}{rgb}{0.21,0.49,0.74}
\usepackage[pagebackref,breaklinks,colorlinks,allcolors=cvprblue]{hyperref}

\def\paperID{26156}
\def\confName{CVPR}
\def\confYear{2026}

\title{\modelname:~\myuline{D}ual stage \myuline{U}nified \myuline{E}fficient \myuline{T}oken reduction for VLM Training and Inference}

\author{Aditya Kumar Singh$^{*}$,\enspace
Hitesh Kandala$^{*}$,\enspace
Pratik Prabhanjan Brahma,\enspace
Zicheng Liu,\enspace
Emad Barsoum\\
Advanced Micro Devices, Inc. (AMD)\\
{\small $^*$Equal contribution}
}

\begin{document}
\maketitle

\begin{abstract}
Vision-language models (VLMs) have achieved remarkable multimodal understanding and reasoning capabilities, yet remain computationally expensive due to dense visual tokenization. 
Existing efficiency approaches either merge redundant visual tokens or drop them progressively in 
language backbone,
often trading accuracy for speed. 
In this work, we propose \modelname, a versatile plug-and-play dual compression framework that consists of (a)~vision-only redundancy aware compression of vision encoder's output into information-preserving tokens,
followed by
(b)~layer-wise, salient text-guided dropping of visual tokens within the language backbone to progressively prune less informative tokens.
This coordinated token management enables aggressive compression while retaining critical semantics. 
On \emph{LLaVA-1.5-7B}, our approach maintains over 99\% of baseline accuracy with 67\% fewer tokens $\downarrow$, and still retains $>$97\% even at 89\% $\downarrow$ reduction. 
With this dual-stage compression during training, it achieves 99.7\% accuracy at 67\% $\downarrow$ and 97.6\% at 89\% $\downarrow$, surpassing prior SoTA visual token reduction methods across multiple benchmarks.
When integrated into \emph{Video-LLaVA-7B}, it even surpasses the baseline--achieving $>$\emph{100\%} $\uparrow$ accuracy with a substantial 53.1\% $\downarrow$ token reduction and retaining \emph{97.6\%} accuracy under an extreme 93.4\% $\downarrow$ setting.
These results highlight end-to-end training with \modelname, enabling robust adaptation to reduced visual (image/video) input without sacrificing accuracy, producing compact yet semantically rich representations within the same computational budget.
Our code is available at \href{https://github.com/AMD-AGI/DUET-VLM}{https://github.com/AMD-AGI/DUET-VLM}.

\end{abstract}

\section{Introduction}
\label{sec:intro}

\begin{figure}[htbp]
  \centering

  \begin{subfigure}{\columnwidth}
    \centering
    \includegraphics[width=0.9\columnwidth]{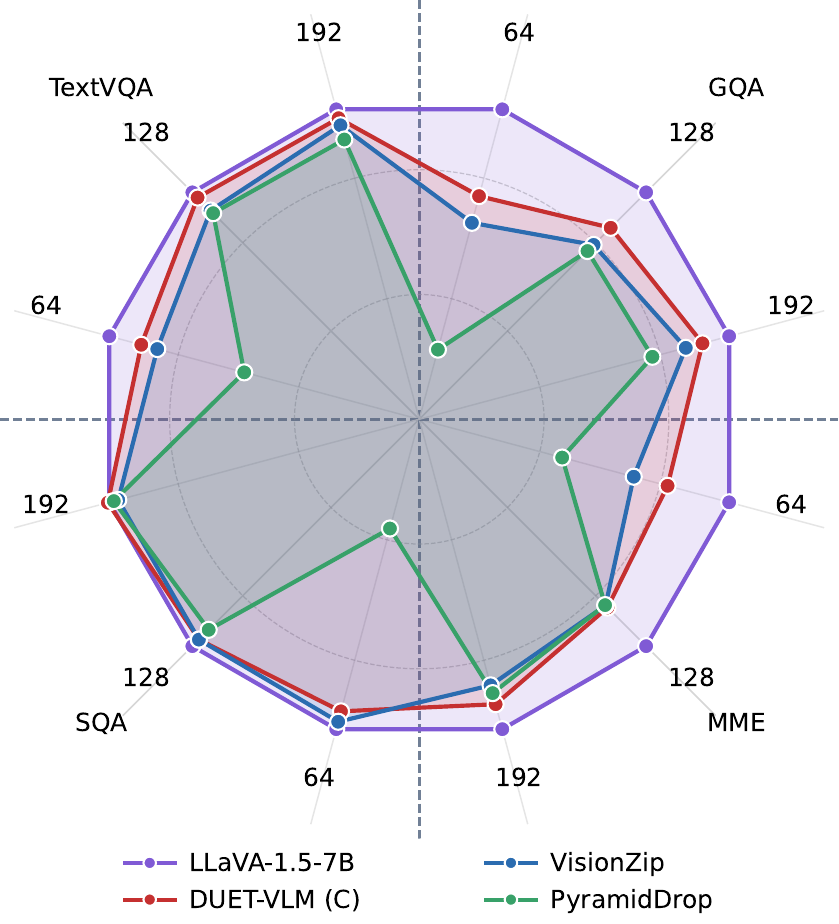} 
    \caption{}
  \end{subfigure}

  \vspace{1ex}

  \begin{subfigure}{\columnwidth}
    \centering
    \includegraphics[width=0.98\columnwidth]{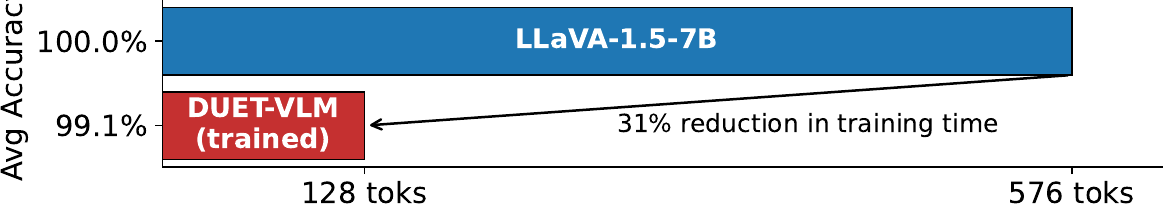}
    \caption{}
  \end{subfigure}
  \caption{\textbf{Efficiency and accuracy comparison of \modelname.} 
    (a) shows the average inference-only accuracy of VisionZip, PyramidDrop, and \modelname across different token budgets, compared to the full 576-token LLaVA-1.5-7B baseline on four benchmarks. 
    (b) demonstrates that trained \modelname model achieves a 31\% reduction in training time while incurring less than a 1\% drop in accuracy relative to the LLaVA-1.5-7B baseline.}
  \label{fig:teaser}
\end{figure}

In the realm of multimodal intelligence, \emph{an image is worth a thousand words} but also thousands of tokens.  
Each image encodes colors, textures, and spatial cues that enrich linguistic reasoning, yet this expressivity burdens vision-language models (VLMs) with an overwhelming number of visual tokens. 
For instance, LLaVA-1.5~\cite{liu2023llava} processes 576 visual tokens per image,
while LLaVA-NeXT~\cite{liu2024llavanext} at $672\times672$ resolution expands this to over 2,800 tokens---orders of magnitude higher than the accompanying text. 
Unlike text, whose token count grows linearly, visual tokens scale \emph{quadratically} with image resolution ($H{\times}W$), compounding the already quadratic attention cost and making large vision-language models (LVLMs) both memory- and latency-bound.

Recent studies~\cite{clip,siglip,liu2023llava} confirm that scaling up visual tokens boosts grounding and reasoning accuracy. Yet, this improvement comes at a high computational cost, motivating a surge of research into efficient visual compression.

Two primary fronts for multimodal efficiency are (i)~\emph{vision-side} and (ii)~\emph{language-side} compression, each targeting redundant visual computation and quadratic attention costs.
(i)~Methods in this category aim to compact the visual embedding space even before cross-modal interaction~\cite{yang2024visionzip,shang2024LLaVA-PruMerge,arif2025hired}.
Whereas
(ii)~prune redundant visual tokens \emph{during or after} their interaction with the language model~\cite{xing2024pyramiddrop,chen2024fastv,ye2025fit}.

\paragraph{Our Perspective.}  
In summary, existing methods either (a) \emph{merge too early}, risking information loss (VisionZip~\cite{yang2024visionzip}, PruMerge~\cite{shang2024LLaVA-PruMerge}), or (b) \emph{drop too uniformly}, lacking semantic adaptivity (PyramidDrop~\cite{xing2024pyramiddrop}, FitPrune~\cite{ye2025fit}). 
Others (FastV~\cite{chen2024fastv}, HiRED~\cite{arif2025hired}) offer partial adaptivity but remain \emph{single directional}, compressing only once, either before or during reasoning, thus failing to jointly optimize \emph{redundancy removal} and \emph{context-aware retention}.

\paragraph{Our Approach.}  
In this paper, we introduce \modelname, a \emph{unified two-way visual compression
framework} that jointly optimizes redundancy removal and contextual adaptivity across the vision-language interface.  
\modelname~operates in two complementary stages:
(1)~\emph{Vision-to-Vision (V2V) merging}, which merges correlated patches into compact embeddings, reducing redundancy before cross-modal fusion; and  
(2)~\emph{Text-to-Vision (T2V) pruning}, which dynamically prunes visual tokens at successive layers using text-to-vision attention scores.  
This dual-stage design, coupling early \emph{structural merging} with late \emph{semantic pruning}, retains essential low-level visual details for grounding early on while progressively removing redundant tokens as reasoning deepens.
As a consequence, it maintains near-baseline accuracy even under extreme token compression, preserving over 99\% of baseline-performance at 67\% token reduction and above 97\% at 89\% reduction during inference across multiple image benchmarks, as shown in \cref{fig:teaser}.
This trend extends to video understanding as well,
where our method surpasses the baseline ($>$100\% accuracy) with 53.1\% token reduction and still achieves 97.6\% accuracy under an extreme \emph{93.4\%} reduction.
These results show that smart token compression can match or even surpass traditional scaling.

\noindent In summary, our contributions are:
\begin{itemize}[leftmargin=*]
    \item \myuline{Joint Vision–Language Optimization}: Unlike prior one-sided compression schemes, we couple both the vision and language backbones through a differentiable compression interface, enabling visual representations and linguistic context to co-adapt during token reduction while maintaining strong cross-modal alignment and efficiency.

    \item \myuline{Dual-Stage Visual Compression}: A unified, task-agnostic framework (\modelname) that first performs \emph{redundancy-aware visual token merging} into compact, information-preserving tokens, followed by \emph{layer-wise saliency-based text-guided token dropping} within language counterpart to prune less informative tokens.
    
    \item \myuline{Accuracy Retention}: \modelname~preserves semantic richness and cross-modal reasoning, maintaining over 99\% of baseline accuracy with only one-third of the original visual tokens across multiple image and video benchmarks.
\end{itemize}
\noindent
Overall, \modelname~shows that careful token management can match model scaling in accuracy while significantly reducing compute.

\section{Related Works}
\label{sec:related_works}

\begin{figure*}[t]
    \centering
    \includegraphics[width=0.98\textwidth]{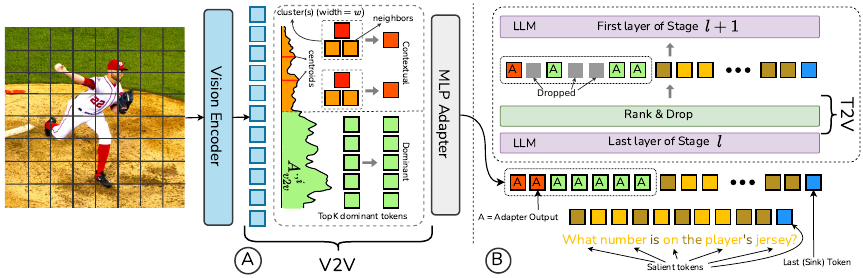}
    \caption{\textbf{Overview of the proposed pipeline.}
    An input image is first encoded into $N$ visual tokens by the \textit{Vision Encoder}.
    (A)~Based on the V2V self-attention map $\avv^{.,i} \coloneqq 1/N\sum_j\avv^{j,i}$, we select the most influential Top-$k_1$ \emph{dominant tokens}, while the remaining tokens, $\Xr$, are merged into $k_2$ \emph{contextual tokens} via attention-guided clustering with a fixed cluster width $w$, to reduce redundancy.
    (B)~The resulting reduced visual tokens, $\Xo$, are fed into a language backbone after projecting it through an MLP \emph{Adapter}, where salient text tokens, $\MS$, further prune visual tokens based on cross-attention scores $\atv$ at certain selected layers called \emph{stages}.}
    \label{fig:method}
\end{figure*}

\paragraph{Token Efficiency in VLMs.}
The rise of LLMs~\cite{openai2023gpt,bai2023qwen,longlora,vicuna2023,zheng2024dape,touvron2023llama,rafailov2023direct} has driven the development of large vision-language models (LVLMs)~\cite{liu2023llava,li2024mgm,tong2025cambrian,instructblip,zhu2023minigpt,chen2023minigptv2}, which align visual encoders such as CLIP~\cite{clip} or SigLIP~\cite{siglip} with powerful text decoders for unified multimodal reasoning.
However, high-resolution vision inputs yield hundreds to thousands of tokens (\eg, $672\times672$ images producing $>2{,}800$ tokens in LLaVA-NeXT~\cite{liu2024llavanext}), causing quadratic growth in transformer attention cost. 
This has led to a spectrum of token-efficiency methods, majorly drawing inspiration from token reduction in LLMs~\cite{kim2022tokenprune,song-etal-2025-less-is-more}.
We broadly categorized these methods as \emph{vision-encoder–side compression} and \emph{language-side compression}.
The former class targets redundancy at the feature-extraction stage: VisionZip~\cite{yang2024visionzip} merges visually similar patches into representative tokens while retaining irreducible dominant ones, HiRED~\cite{arif2025hired} uses CLS-guided saliency to allocate a fixed token budget across high-resolution partitions, and PruMerge~\cite{shang2024LLaVA-PruMerge} merges unpruned tokens via attention-sparsity clustering.
These approaches effectively reduce visual clutter before multimodal fusion, but their static or heuristic merging strategies limit downstream adaptivity. 
In contrast, language-side compression techniques such as PyramidDrop~\cite{xing2024pyramiddrop} progressively drop visual tokens across transformer layers, FastV~\cite{chen2024fastv} learns adaptive attention masks to prune tokens in deeper stages, and FitPrune~\cite{ye2025fit} devises a training-free pruning recipe matching pre- and post-pruning attention distributions. 
While these methods achieve substantial FLOP savings, they operate within a single stage either compressing early or pruning late thus lacking synergy between the vision and language representations.

\paragraph{Joint Multimodal Compression.}
Recent studies suggest that redundancy evolves across both spatial and semantic hierarchies~\cite{xing2024pyramiddrop, yang2024visionzip}, motivating approaches that reason jointly over vision and text-conditioned saliency. 
Our work builds upon this insight, extending VisionZip’s early-stage merging and PyramidDrop’s progressive pruning under a unified, differentiable framework. 
Unlike these prior methods, we merge fewer yet contextually coherent tokens per cluster in the vision encoder, preserving structural diversity, and incorporate salient text-guided tokens for layer-wise adaptive dropping in the LLM. 
This coupling of vision-side compaction with language-guided pruning achieves joint optimization under a fixed token budget, yielding higher accuracy than existing single-stage compressors across multiple multimodal benchmarks. 
A detailed formulation of this dual-stage pipeline linking redundancy-aware merging and saliency-driven pruning is presented next in our \cref{sec:method}.

\section{Methodology}
\label{sec:method}
Our approach primarily combines VisionZip's~\cite{yang2024visionzip} visual token reduction 
followed by PyramidDrop's~\cite{xing2024pyramiddrop} hierarchical text-based visual tokens pruning.
VisionZip reduces all visual tokens to (a)~dominant tokens and (b)~contextual tokens, where we modify contextual token construction via our novel approach of local cluster aggregation as formalised in~\cref{subsec:clustering,subsubsec:motiv_local} and \cref{fig:method}(A).
For PyramidDrop, instead of last text token, we experiment with various salient token selection methods
as shown in~\cref{subsec:vision_token_dropping} and further experimented in \cref{tab:token_sel_llava,tab:token_sel_llava_next}.

\subsection{Clustering of vision tokens}
\label{subsec:clustering}
Let 
\(
\bX \coloneqq \{\bx_i\}_{i=1}^{N} \in \bbR^{N \times d}
\)
denote the set of $N$ vision tokens
from the vision encoder,
with $ d $ being the embedding dimension.
With help of self-attention map $ \avv \in \bbR^{N \times N} $ retrieved from CLIP's last layer, we further identify a subset of $k_1$ dominant tokens, $\Xd$, that has high vision-to-vision affinity,
where
$|\Xd| = k_1 < N$.
For residual vision tokens, we merge them into $k_2$ tokens via clustering, as shown in \cref{alg:dominant-residual}.



\makeatletter
\newcommand{\StateSpace}{\vspace{4pt}} 
\makeatother

\begin{algorithm}[t]
\caption{Attention guided Dominant selection + Residual Clustering}
\label{alg:dominant-residual}
\begin{algorithmic}[1]
\Require Tokens $\bX=\{\bx_i\}_{i=1}^N$, attention map $\avv\in\mathbb{R}^{N\times N}$, $k_1,k_2,w$ ( = cluster-size)
\Ensure Reduced tokens $\Xo = \Xd \cup \Xc$, where $\Xc =$ contextual tokens

\State Functions: $ \topk() $ takes two arguments, (i)~an array and (ii)~size of set of top elements in the array to be selected. 
\StateSpace
\State Compute attention score $s_i=\sum_{j=1}^N \avv^{j,i}$, select dominant indices $\MD \coloneqq \topk(\{s_i\},k_1)$ and residual indices $\MR\coloneqq\{1,\dots,N\}\setminus\MD$.
\StateSpace
\State From $\MR$, pick cluster centroids 
{
\setlength{\abovedisplayskip}{3pt} 
\setlength{\belowdisplayskip}{3pt} 
\begin{equation}
    \MC\coloneqq\topk(\{s_i\}_{i\in\MR},k_2)
\end{equation}
}
\State For each centroid $c\in\MC$, get neighbors' indices
{
\setlength{\abovedisplayskip}{3pt} 
\setlength{\belowdisplayskip}{3pt} 
\begin{equation}
\MN_c=\topk(\avv^{c,\MR},w)    
\end{equation}
}
and form $ \bz_c=\sum_{j\in\MN_c}\bx_j/|\MN_c|$.
\StateSpace
\State Return $\Xd=\{\bx_i\}_{i\in\MD}$ and $\Xc=\{\bz_c\}_{c\in\MC}$, \ie  $\Xo=\Xd\cup\Xc$.
\end{algorithmic}
\end{algorithm}
\subsubsection{Motivation for local clustering}
\label{subsubsec:motiv_local}
\paragraph{a.~Semantic misalignment.}~The dominant tokens
often correspond to high-activation visual areas which might not be semantically relevant for downstream text-guided tasks.

\paragraph{b.~Information dilution.}~When the volume of residual tokens is large, \ie
$ |\Xr| \gg k_2 $, where $\Xr \coloneqq \bX \setminus \Xd $ all visual tokens except $k_1$ dominant tokens, averaging
dilutes the semantically informative but low-frequency features.

\paragraph{Local Cluster Aggregation.}
To address these limitations, we replace global averaging with \emph{local cluster aggregation}. 
Instead of merging all non-dominant tokens into a few global contextual vectors, we partition the residual tokens $\Xr$ into $k_2$ small
clusters with centroid $c$ and their neighbors $\{\MN_c\}_{c\in \MC}$
where each cluster produces one contextual token via local averaging.
We choose centroids based on the ranking of V2V self-attention scores ($\avv$) to have a good basis for forming the context.
Importantly, we do \emph{not} increase the number of contextual tokens; instead, we maintain 
$k_2$, while enforcing a small cluster width $w$, where $w = |\MN_c|$. 
This has two advantages:
(a)~\emph{Aggregation} is performed within a local neighborhood $\MN_c$, preserving fine-grained cues while avoiding dilution by background noise. 
(b)~Since $w.k_2 < |\Xr|$, the unassigned tokens in $\Xr$ are \emph{dropped} early, effectively removed from the pipeline before entering the language backbone.
For example, with $N{=}576$, $k_1{=}54$, $k_2{=}10$, and $w{=}4$, a total of $576 - 54 - (10 {\cdot} 4) = 482$ tokens are dropped before reaching the language backbone.

\subsection{Hierarchical visual tokens dropping}
\label{subsec:vision_token_dropping}

For LLM, we wrap the compact set of visual tokens with textual tokens, \ie, a \emph{system prompt} at the beginning and \emph{query} tokens at the end.

Let $\MT \coloneqq \{t_k\}_{k=1}^m$ be the full sequence of text tokens (queries).
Importantly, while the entire text sequence $\MT$ is preserved and passed through the network, only a subset of salient text tokens, $\MS$ identified by the salient-token estimator $F_\MS$(.), is used to compute text-to-vision (T2V) saliency, which then forms the basis for progressive visual token pruning across the language layers.
The last text token ($t_m$) is always included in $\MS$ as a \emph{sink} token
to stabilize the attention and
capture globally relevant prompt-context as well as visually critical regions even under aggressive pruning~\cite{xing2024pyramiddrop}.

\paragraph{Text-Guided Visual Token Dropping.}
Let us assume we have $L$ layers in the language backbone, grouped into $M$ stages with $M<L$.
We define $\MVl$ to be set of visual tokens retained at stage $l\in\{1,\dots,M\}$ with $N_l \coloneqq |\MVl|$.
And for $l=0$, we have $\MV^{(0)} = \Xo$ and $N_0 = |\Xo| = k_1 + k_2$ visual tokens.
Now, given a target compression ratio $\lambda$, after each stage $l$, we drop $\lfloor\lambda.N_l\rfloor$ or retain $\lfloor(1-\lambda).N_l\rfloor$ tokens based on T2V cross-attention scores $\atv^{(l)}\in \bbR^{|\MS| \times N_l}$ between the salient text tokens $\MS$ and retained vision tokens $\MVl$, as shown in~\cref{fig:method}(B).
First, we rank visual tokens ($\MVl$) in descending order and then we choose first $\lfloor(1-\lambda).N_l\rfloor$ tokens, essentially doing $\topk(\MVl, \lfloor(1-\lambda).N_l\rfloor)$.
This rank-and-drop procedure is repeated at each stage, progressively narrowing the visual context by filtering redundant visual information as multimodal reasoning deepens.
This enforces an adaptive balance between computational efficiency and representational fidelity and ensures that higher layers attend only to semantically relevant regions.
\section{Experiment}
\label{sec:expts}





\subsection{Setup}
\label{subsec:setup}

\begin{table*}[ht]
\centering
\tabcolsep=0.1cm
\caption{
\textbf{Comparison of inference-only methods on LLaVA-1.5-7B.} 
We report results across five benchmarks under different average token budgets, corresponding to 67\%, 78\%, and 89\% token reduction. 
\modelname~\crm achieves the highest average performance across all settings, maintaining over 99\% of the baseline accuracy while using significantly fewer tokens. 
}
\vspace{-2mm}
\label{tab:final_inference_table}
\begin{tabular}{l ccccc c }
\toprule
\multicolumn{1}{l|}{Method} &
  POPE &
  SQA\textsuperscript{I} &
  VQA\textsuperscript{T} &
  MME &
  \multicolumn{1}{c|}{GQA} &
  Avg \\ \midrule
\rowcolor{gray!20}
\multicolumn{7}{c}{Total 576 Tokens (100.0\%)} \\ 
\multicolumn{1}{l|}{LLaVA-1.5-7B} &
  85.9 (100.0\%) &
  69.5 (100.0\%) &
  58.2 (100.0\%) &
  1862 (100.0\%) &
  \multicolumn{1}{c|}{61.9 (100.0\%)} &
  100.0\% \\ \midrule
\rowcolor{gray!20}
\multicolumn{7}{c}{Avg 192 Tokens (~\textcolor{red}{\textdownarrow~66.7\%})} \\ 
\multicolumn{1}{l|}{FastV~\cite{chen2024fastv} {[}ECCV’24{]}} &
  64.8 (75.4\%) &
  67.3 (96.8\%) &
  52.5 (90.2\%) &
  1612 (86.6\%) &
  \multicolumn{1}{c|}{52.7 (85.1\%)} &
  86.8\% \\
\multicolumn{1}{l|}{HiRED~\cite{arif2025hired} {[}AAAI’25{]}} &
  82.8 (96.4\%) &
  68.4 (98.4\%) &
  47.4 (81.4\%) &
  1737 (93.3\%) &
  \multicolumn{1}{c|}{58.7 (94.8\%)} &
  92.9\% \\
\multicolumn{1}{l|}{FitPrune~\cite{ye2025fit} {[}AAAI’25{]}} &
  83.4 (97.1\%) &
  67.8 (97.6\%) &
  {{\ul 57.4} ({\ul 98.6\%})} &
  \textbf{1831 (98.3\%)} &
  \multicolumn{1}{c|}{\textbf{60.4 (97.6\%)}} &
  {\ul 97.8\%} \\
\multicolumn{1}{l|}{PruMerge~\cite{shang2024LLaVA-PruMerge} {[}ICCV’25{]}} &
  71.3 (83.0\%) &
  67.9 (97.7\%) &
  54.3 (93.3\%) &
  1632 (87.6\%) &
  \multicolumn{1}{c|}{54.3 (87.7\%)} &
  89.9\% \\
\multicolumn{1}{l|}{SparseVLM~\cite{zhang2024sparsevlm} {[}ICML’25{]}} &
  83.6 (97.3\%) &
  69.1 (99.4\%) &
  56.1 (96.4\%) &
  1721 (92.4\%) &
  \multicolumn{1}{c|}{57.6 (93.1\%)} &
  95.7\% \\
\multicolumn{1}{l|}{PyramidDrop~\cite{xing2024pyramiddrop} {[}CVPR’25{]}} &
  - (-) &
  {{\ul 69.2} ({\ul 99.6\%})} &
  56.5 (97.1\%) &
  1797 (96.5\%) &
  \multicolumn{1}{c|}{57.3 (92.6\%)} &
  96.4\% \\
\multicolumn{1}{l|}{VisionZip~\cite{yang2024visionzip} {[}CVPR’25{]}} &
  {{\ul 85.3} ({\ul 99.3\%})} &
  68.9 (99.1\%) &
  57.3 (98.5\%) &
  1783 (95.8\%) &
  \multicolumn{1}{c|}{59.3 (95.8\%)} &
  97.7\% \\
\rowcolor{lightyellow}
\multicolumn{1}{l|}{\modelname~\crm} &
  \textbf{86.5 (100.7\%)} &
  \textbf{69.6 (100.1\%)} &
  \textbf{57.7 (99.1\%)} &
  {{\ul 1817} ({\ul 97.6\%})} &
  \multicolumn{1}{c|}{{{\ul 60.3} ({\ul 97.4\%})}} &
  \textbf{99.0\%} \\ \midrule
\rowcolor{gray!20}
\multicolumn{7}{c}{Avg 128 Tokens (~\textcolor{red}{\textdownarrow~77.8\%})} \\ 
\multicolumn{1}{l|}{FastV~\cite{chen2024fastv}{[}ECCV’24{]}} &
  59.6 (69.4\%) &
  60.2 (86.6\%) &
  52.5 (90.2\%) &
  1490 (80.0\%) &
  \multicolumn{1}{c|}{49.6 (80.1\%)} &
  81.3\% \\
\multicolumn{1}{l|}{HiRED~\cite{arif2025hired}{[}AAAI’25{]}} &
  79.8 (92.9\%) &
  68.1 (98.0\%) &
  46.1 (79.2\%) &
  1710 (91.8\%) &
  \multicolumn{1}{c|}{57.2 (92.4\%)} &
  90.9\% \\
\multicolumn{1}{l|}{FitPrune~\cite{ye2025fit}{[}AAAI’25{]}} &
  77.9 (90.7\%) &
  68.0 (97.8\%) &
  55.7 (95.7\%) &
  \textbf{1776 (95.4\%)} &
  \multicolumn{1}{c|}{{{\ul 58.5} ({\ul 94.5\%})}} &
  94.8\% \\
\multicolumn{1}{l|}{PruMerge~\cite{shang2024LLaVA-PruMerge} {[}ICCV’25{]}} &
  67.2 (78.2\%) &
  67.1 (96.5\%) &
  54.3 (93.3\%) &
  1554 (83.5\%) &
  \multicolumn{1}{c|}{53.3 (86.1\%)} &
  87.5\% \\
\multicolumn{1}{l|}{SparseVLM~\cite{zhang2024sparsevlm} {[}ICML’25{]}} &
  80.5 (93.7\%) &
  68.4 (98.4\%) &
  54.9 (94.3\%) &
  1696 (91.1\%) &
  \multicolumn{1}{c|}{56.0 (90.5\%)} &
  93.6\% \\
\multicolumn{1}{l|}{PyramidDrop~\cite{xing2024pyramiddrop} {[}CVPR’25{]}} &
  - (-) &
  68.4 (98.4\%) &
  56.6 (97.3\%) &
  1761 (94.6\%) &
  \multicolumn{1}{c|}{57.1 (92.2\%)} &
  95.6\% \\
\multicolumn{1}{l|}{VisionZip~\cite{yang2024visionzip} {[}CVPR’25{]}} &
  {{\ul 83.2} ({\ul 96.9\%})} &
  {{\ul 68.9} ({\ul 99.1\%})} &
  {{\ul 56.8} ({\ul 97.6\%})} &
  1762 (94.6\%) &
  \multicolumn{1}{c|}{57.6 (93.1\%)} &
  {\ul 96.3\%} \\
\rowcolor{lightyellow}
\multicolumn{1}{l|}{\modelname~\crm} &
  \textbf{85.9 (100.0\%)} &
  \textbf{70.2 (101.0\%)} &
  \textbf{57.8 (99.3\%)} &
  {{\ul 1767} ({\ul 94.9\%})} &
  \multicolumn{1}{c|}{\textbf{59.0 (95.3\%)}} &
  \textbf{98.1\%} \\ \midrule
\rowcolor{gray!20}
\multicolumn{7}{c}{Avg 64 Tokens (~\textcolor{red}{\textdownarrow~88.9\%})} \\
\multicolumn{1}{l|}{FastV~\cite{chen2024fastv}{[}ECCV’24{]}} &
  48.0 (55.9\%) &
  51.1 (73.5\%) &
  47.8 (82.1\%) &
  1256 (67.5\%) &
  \multicolumn{1}{c|}{46.1 (74.5\%)} &
  70.7\% \\
\multicolumn{1}{l|}{HiRED~\cite{arif2025hired}{[}AAAI’25{]}} &
  73.6 (85.7\%) &
  68.2 (98.1\%) &
  44.2 (75.9\%) &
  1599 (85.9\%) &
  \multicolumn{1}{c|}{54.6 (88.2\%)} &
  86.8\% \\
\multicolumn{1}{l|}{FitPrune~\cite{ye2025fit}{[}AAAI’25{]}} &
  60.9 (70.9\%) &
  68.0 (97.8\%) &
  51.2 (88.0\%) &
  1556 (83.6\%) &
  \multicolumn{1}{c|}{52.3 (84.5\%)} &
  85.0\% \\
\multicolumn{1}{l|}{PruMerge~\cite{shang2024LLaVA-PruMerge} {[}ICCV’25{]}} &
  65.3 (76.0\%) &
  68.1 (98.0\%) &
  54.0 (92.8\%) &
  1549 (83.2\%) &
  \multicolumn{1}{c|}{51.9 (83.8\%)} &
  86.8\% \\
\multicolumn{1}{l|}{SparseVLM~\cite{zhang2024sparsevlm} {[}ICML’25{]}} &
  75.1 (87.4\%) &
  62.2 (89.5\%) &
  51.8 (89.0\%) &
  1505 (80.8\%) &
  \multicolumn{1}{c|}{52.7 (85.1\%)} &
  86.4\% \\
\multicolumn{1}{l|}{PyramidDrop~\cite{xing2024pyramiddrop} {[}CVPR’25{]}} &
  - (-) &
  \textbf{69.0 (99.3\%)} &
  50.6 (86.9\%) &
  1561 (83.8\%) &
  \multicolumn{1}{c|}{47.5 (76.7\%)} &
  86.7\% \\
\multicolumn{1}{l|}{VisionZip~\cite{yang2024visionzip} {[}CVPR’25{]}} &
  77.0 (89.6\%) &
  \textbf{69.0 (99.3\%)} &
  {{\ul 55.5} ({\ul 95.4\%})} &
  {{\ul 1690} ({\ul 90.8\%})} &
  \multicolumn{1}{c|}{{{\ul 55.1} ({\ul 89.0\%})}} &
  {\ul 92.8\%} \\
\rowcolor{lightyellow}
\multicolumn{1}{l|}{\modelname~\crm} &
  \textbf{82.5 (96.0\%)} &
  {{\ul 68.3} ({\ul 98.3\%})} &
  \textbf{56.4 (96.9\%)} &
  \textbf{1751 (94.0\%)} &
  \multicolumn{1}{c|}{\textbf{56.7 (91.6\%)}} &
  \textbf{95.4\%} \\ \bottomrule
\end{tabular}%
\vspace{-3mm}
\end{table*}

\begin{table*}[t]
\centering
\tabcolsep=0.1cm
\caption{\textbf{Comparison of inference-only methods on Qwen-2.5-VL-7B}. We report results across five benchmarks under different average token budgets. \modelname~\crm achieves the highest average performance across all settings, maintaining over 98\% of the baseline accuracy while using significantly fewer tokens. }
\begin{tabular}{l|c|ccccc|c}
\toprule
\multicolumn{1}{l|}{Method} & \multicolumn{1}{c|}{Speedup} & POPE & SQA\textsuperscript{I} & VQA\textsuperscript{T} & MME & \multicolumn{1}{c|}{GQA} & Avg \\
\midrule
\rowcolor{gray!20}
\multicolumn{8}{c}{Dynamic Tokens}\\
\multicolumn{1}{l|}{Qwen-2.5-VL-7B~\cite{bai2025qwen25vltechnicalreport}} & \multicolumn{1}{c|}{1.00$\times$} & 87.8 (100.0\%) & 86.2 (100.0\%) & 72.3 (100.0\%) & 2264 (100.0\%) & \multicolumn{1}{c|}{58.3 (100.0\%)} & 100.0\% \\
\midrule
\rowcolor{gray!20}
\multicolumn{8}{c}{Avg 640 Tokens}\\
\multicolumn{1}{l|}{VisionZip} & \multicolumn{1}{c|}{1.3x} & 87.8 (100.0\%) & \textbf{86.2 (100.0\%)} & 71.2 (98.5\%) & \textbf{2267 (100.1\%)} & \multicolumn{1}{c|}{58.3 (100.0\%)} & 99.7\% \\
\rowcolor{lightyellow}
\multicolumn{1}{l|}{DUET-VLM (C)} & \multicolumn{1}{c|}{1.3x} & \textbf{88.3 (100.6\%)} & 85.8 (99.5\%) & \textbf{71.5 (98.9\%)} & 2264 (100.0\%) & \multicolumn{1}{c|}{\textbf{58.5 (100.3\%)}} & \textbf{99.9\%} \\
\midrule
\rowcolor{gray!20}
\multicolumn{8}{c}{Avg 320 Tokens}\\
\multicolumn{1}{l|}{VisionZip} & \multicolumn{1}{c|}{1.4x} & 88.0 (100.2\%) & \textbf{86.2 (100.0\%)} & 69.4 (96.0\%) & 2251 (99.4\%) & \multicolumn{1}{c|}{58.2 (99.8\%)} & 99.1\% \\
\rowcolor{lightyellow}
\multicolumn{1}{l|}{DUET-VLM (C)} & \multicolumn{1}{c|}{1.4x} & \textbf{88.3 (100.6\%)} & 85.8 (99.5\%) & \textbf{71.3 (98.6\%)} & \textbf{2259 (99.8\%)} & \multicolumn{1}{c|}{\textbf{58.5 (100.3\%)}} & \textbf{99.8\%} \\
\midrule
\rowcolor{gray!20}
\multicolumn{8}{c}{Avg 160 Tokens}\\
\multicolumn{1}{l|}{VisionZip} & \multicolumn{1}{c|}{1.5x} & 87.4 (99.5\%) & \textbf{86.1 (99.9\%)} & 64.6 (89.3\%) & 2210 (97.6\%) & \multicolumn{1}{c|}{57.3 (98.3\%)} & 96.9\% \\
\rowcolor{lightyellow}
\multicolumn{1}{l|}{DUET-VLM (C)} & \multicolumn{1}{c|}{1.5x} & \textbf{88.2 (100.5\%)} & 86.0 (99.8\%) & \textbf{67.7 (93.6\%)} & \textbf{2235 (98.7\%)} & \multicolumn{1}{c|}{\textbf{57.9 (99.3\%)}} & \textbf{98.4\%} \\
\bottomrule
\end{tabular}
\label{tab:rebuttal_pope_vqa_tokenbudgets}
\end{table*}
\begin{table*}[t]
\centering
\tabcolsep=0.1cm
\caption{\textbf{Comparison of trained methods on LLaVA-1.5-7B.} We report scores across multiple benchmarks under different visual token budgets. DUET-VLM variants (C), (C+all), and (C+S) defined in~\cref{subsec:setup} consistently outperform or match prior compression approaches such as PyramidDrop and VisionZip, even with reductions of up to 88.9\% in visual tokens. These results highlight the effectiveness of training of our dual-stage token selection strategy in preserving baseline accuracy under aggressive compression.
}
\label{tab:train_ablation}
\begin{tabular}{l ccccc c}
\toprule
\multicolumn{1}{l|}{Method} &
  POPE &
  SQA\textsuperscript{I} &
  VQA\textsuperscript{T} &
  MME &
  \multicolumn{1}{c|}{GQA} &
  Avg \\ \midrule
\rowcolor{gray!20}
\multicolumn{7}{c}{Total 576 Tokens} \\
\multicolumn{1}{l|}{LLaVA-1.5-7B~\cite{llava2024}} &
  85.9 (100.0\%) &
  69.5 (100.0\%) &
  58.2 (100.0\%) &
  1862.0 (100.0\%) &
  \multicolumn{1}{c|}{61.9 (100.0\%)} &
  100.0\% \\ \midrule
\rowcolor{gray!20}
\multicolumn{7}{c}{Avg 192 Tokens (~\textcolor{red}{\textdownarrow~66.7\%})} \\
\multicolumn{1}{l|}{VisionZip} &
  84.9 (98.8\%) &
  68.2 (98.1\%) &
  57.8 (99.3\%) &
  \textbf{1834 (98.5\%)} &
  \multicolumn{1}{c|}{{\ul 60.1} ({\ul 97.1\%})} &
  {\ul 98.4\%} \\
\rowcolor{lg}
\multicolumn{1}{l|}{\modelname~\crm} &
  \textbf{85.5 (99.5\%)} &
  {\ul 69.6} ({\ul 100.1\%}) &
  {\ul 58.2} ({\ul 100.0\%}) &
  {\ul 1826} ({\ul 98.1\%}) &
  \multicolumn{1}{c|}{\textbf{62.4 (100.8\%)}} &
  \textbf{99.7\%} \\
\rowcolor{lightpink}
\multicolumn{1}{l|}{\modelname~\call} &
  {\ul 85.4} ({\ul 99.4\%}) &
  {\ul 69.6} ({\ul 100.1\%}) &
  57.7 (99.1\%) &
  1790 (96.1\%) &
  \multicolumn{1}{c|}{57.6 (93.1\%)} &
  97.6\% \\
\rowcolor{lightyellow}
\multicolumn{1}{l|}{\modelname~\cs} &
  \textbf{85.5 (99.5\%)} &
  \textbf{69.8 (100.4\%)} &
  \textbf{58.5 (100.5\%)} &
  1820 (97.7\%) &
  \multicolumn{1}{c|}{57.6 (93.1\%)} &
  98.3\% \\ \midrule
\rowcolor{gray!20}
\multicolumn{7}{c}{Avg 128 Tokens (~\textcolor{red}{\textdownarrow~77.8\%})} \\
\multicolumn{1}{l|}{VisionZip} &
  83.4 (97.1\%) &
  {\ul 68.5} ({\ul 98.6\%}) &
  57.1 (98.1\%) &
  1748 (93.9\%) &
  \multicolumn{1}{c|}{57.6 (93.1\%)} &
  96.1\% \\
\rowcolor{lg}
\multicolumn{1}{l|}{\modelname~\crm} &
  {\ul 85.1} ({\ul 99.1\%}) &
  {\ul 68.5} ({\ul 98.6\%}) &
  \textbf{58.4 (100.3\%)} &
  \textbf{1817 (97.6\%)} &
  \multicolumn{1}{c|}{{\ul 61.9} ({\ul 100.0\%})} &
  \textbf{99.1\%} \\
\rowcolor{lightpink}
\multicolumn{1}{l|}{\modelname~\call} &
  \textbf{85.4 (99.4\%)} &
  67.6 (97.3\%) &
  \textbf{58.4 (100.3\%)} &
  1760 (94.5\%) &
  \multicolumn{1}{c|}{\textbf{60.2 (97.3\%)}} &
  97.8\% \\
\rowcolor{lightyellow}
\multicolumn{1}{l|}{\modelname~\cs} &
  {\ul 85.1} ({\ul 99.1\%}) &
  \textbf{69.3 (99.7\%)} &
  {\ul 58.1} ({\ul 99.8\%}) &
  {\ul 1815} ({\ul 97.5\%}) &
  \multicolumn{1}{c|}{60.1 (97.1\%)} &
  {\ul 98.6\%} \\ \midrule
\rowcolor{gray!20}
\multicolumn{7}{c}{Avg 64 Tokens (~\textcolor{red}{\textdownarrow~88.9\%})} \\
\multicolumn{1}{l|}{VisionZip} &
  80.9 (94.2\%) &
  68.8 (99.0\%) &
  56.0 (96.2\%) &
  \textbf{1756 (94.3\%)} &
  \multicolumn{1}{c|}{57.0 (92.1\%)} &
  95.2\% \\
\rowcolor{lg}
\multicolumn{1}{l|}{\modelname~\crm} &
  {\ul 83.7} ({\ul 97.4\%}) &
  {\ul 69.5} ({\ul 100.0\%}) &
  {\ul 57.2} ({\ul 98.3\%}) &
  1737 (93.3\%) &
  \multicolumn{1}{c|}{{\ul 60.0} ({\ul 96.9\%})} &
  {\ul 97.2\%} \\
\rowcolor{lightpink}
\multicolumn{1}{l|}{\modelname~\call} &
  \textbf{83.8 (97.6\%)} &
  68.0 (97.8\%) &
  \textbf{57.6 (99.0\%)} &
  1736 (93.2\%) &
  \multicolumn{1}{c|}{57.6 (93.1\%)} &
  96.1\% \\
\rowcolor{lightyellow}
\multicolumn{1}{l|}{\modelname~\cs} &
  81.9 (95.3\%) &
  \textbf{69.8 (100.4\%)} &
  57.1 (98.1\%) &
  {\ul 1754} ({\ul 94.2\%}) &
  \multicolumn{1}{c|}{\textbf{61.8 (99.8\%)}} &
  \textbf{97.6\%} \\ \bottomrule
\end{tabular}
\end{table*}

\begin{table}[!ht]
\centering
\tabcolsep=0.10cm
\caption{\textbf{Comparison of different methods on Video-LLaVA-7B.} We evaluate performance across three benchmarks to demonstrate that \modelname~preserves accuracy even under substantial token compression.}
\vspace{-1mm}
\label{tab:video-llava}
\begin{tabular}{l ccc c}
\toprule
\multicolumn{1}{l|}{Method}      & TGIF & MSVD & \multicolumn{1}{c|}{MSRVTT} & Avg     \\ \midrule
\rowcolor{gray!20}
\multicolumn{5}{c}{Upper Bound, 2048 Tokens (100\%)}                                   \\ \midrule
\multicolumn{1}{l|}{Video-LLaVA} & 47.1 & 69.8 & \multicolumn{1}{c|}{56.7}   & 100.0\% \\ \midrule
\rowcolor{gray!20}
\multicolumn{5}{c}{Avg 960 Tokens (~\textcolor{red}{\textdownarrow~53.12\%})}                           \\
\multicolumn{1}{l|}{PyramidDrop} & 46.9 & 70.0 & \multicolumn{1}{c|}{58.0}   & 100.7\% \\
\rowcolor{lightyellow}
\multicolumn{1}{l|}{\modelname~\crm} & 48.9 & 70.1 & \multicolumn{1}{c|}{55.6} & \textbf{100.8\%} \\ \midrule
\rowcolor{gray!20}
\multicolumn{5}{c}{Avg 136 Tokens (~\textcolor{red}{\textdownarrow~93.4\%})}                            \\
\multicolumn{1}{l|}{FastV}       & 23.1 & 38.0 & \multicolumn{1}{c|}{19.3}   & 45.8\%  \\
\multicolumn{1}{l|}{SparseVLM}   & 44.7 & 68.2 & \multicolumn{1}{c|}{31.0}   & 82.4\%  \\
\multicolumn{1}{l|}{VisionZip}   & 42.4 & 63.5 & \multicolumn{1}{c|}{52.1}   & 91.0\%  \\
\rowcolor{lightyellow}
\multicolumn{1}{l|}{\modelname~\crm} & 46.3 & 67.7 & \multicolumn{1}{c|}{55.2} & \textbf{97.6\%}  \\ \bottomrule
\end{tabular}
\vspace{-3mm}
\end{table}

\begin{table}[t]
\centering
\caption{\textbf{Inference-only speedup comparison on LLaVA-1.5-7B.} DUET-VLM (C) achieves a strong accuracy-latency trade-off under reduced token budgets.}
\label{tab:inference_only_no_latency}
\begin{tabular}{l|ccc}
\toprule
Method & Tokens & Speedup & Avg \\
\midrule
LLaVA-1.5-7B   & 576 (100\%) & 1x    & 100.0\% \\ \midrule
VisionZip  & 192 (33\%) & 1.1x & 97.7\% \\
PyramidDrop& 192 (33\%) & 1x & 96.4\% \\
DUET-VLM (C)   & 192 (33\%) & 1.1x & 99.0\% \\ \midrule
DUET-VLM (C)   & 64 (11\%)  & 1.2x & 95.4\% \\
\bottomrule
\end{tabular}
\end{table}
\begin{table}[t]
\centering
\caption{\textbf{Training time comparison on LLaVA-1.5-7B.} DUET-VLM (C) offers favorable accuracy--efficiency trade-offs with over 30\% reduction in training time while still maintaining over 99\% accuracy.}
\label{tab:trained_method_comparisons}
\begin{tabular}{l|ccc}
\toprule
Method & Tokens & Train Time $\downarrow$ & Avg \\
\midrule
LLaVA-1.5-7B & 576 & 0\%  & 100.0\% \\ \midrule
DUET-VLM (C) & 192 & 26\% & 99.7\% \\
DUET-VLM (C) & 128 & 31\% & 99.1\% \\
DUET-VLM (C) & 64  & 36\% & 95.4\% \\
\bottomrule
\end{tabular}
\end{table}
\textbf{Implementation Details.} We evaluate our method across three settings. 
For image-based tasks, we use LLaVA-1.5-7B~\cite{llava2024} with 576 vision tokens from the vision encoder. 
For inference-only comparisons, we include LLaVA-NeXT-7B~\cite{liu2024llavanext}, which supports 2,880 vision tokens. 
Training experiments are conducted on LLaVA-1.5-7B, while for video understanding tasks we use Video-LLaVA-7B~\cite{lin2023videollava}, which processes 8 frames per video with up to 2,048 vision tokens.
Unless otherwise specified, we adopt a default cluster width of 4 and an LLM configuration 
\{\emph{layer}: \emph{token-retention-ratio}\} $= \{16{:}0.5, 24{:}0\}$, 
indicating that 50\% of the tokens are retained after the 16\textsuperscript{th} layer, while all remaining tokens are dropped after the 24\textsuperscript{th} layer. 
This configuration is chosen to balance mid-layer contextualization with late-layer sparsity, consistent with prior findings~\cite{xing2024pyramiddrop} that deeper layers contribute less to token-level reasoning.
Consistent with these findings, our experiments show that pruning tokens entirely in the final layers (\eg, after the 24\textsuperscript{th}) yields performance nearly identical to retaining them, 
suggesting that late-stage pruning does not harm multimodal understanding. 
Please refer to the supplemental for detailed ablations supporting this choice. 
All experiments were conducted on \emph{one node} of a compute cluster equipped with 8~$\times$~\AMDGPU~MI325 GPUs. 
We evaluate four variants of our model throughout the paper:
\begin{itemize}[leftmargin=*]
    \item \myuline{\modelname~(vanilla):} Baseline combining VisionZip-style clustering with the original PyramidDrop strategy, where the \emph{last text token} is used to rank and prune visual tokens within the LLM stages.

    \item \myuline{\modelname~\crm:} Variant employing our \emph{local clustering} mechanism in place of VisionZip’s method, still guided by the \emph{last text token} for visual token pruning.

    \item \myuline{\modelname~\call:} Extension using local clustering with attention-guided pruning from \emph{all text tokens} for ranking and dropping visual tokens.

    \item \myuline{\modelname~\cs:} Full model variant combining local clustering with again attention-guided pruning but by the \emph{salient text tokens}, yielding the strongest semantic alignment between vision and language.
\end{itemize}




\paragraph{Benchmarks.} We conduct extensive evaluations on both image and video understanding benchmarks to assess the effectiveness of our approach. 
For image-based tasks, we benchmark our model on POPE~\cite{li-etal-2023-evaluating-pope}, GQA~\cite{hudson2019gqa}, TextVQA (VQA\textsuperscript{T})~\cite{singh2019towards}, MME~\cite{yin2024survey}, SQA-Image (SQA\textsuperscript{I})~\cite{lu2022learn}, and SeedBench-Image (Seed\textsuperscript{I})~\cite{li2023seed}. 
For video, we follow the evaluation protocol proposed in Video-LLaVA~\cite{lin2023videollava} 
and evaluate on three widely used video question answering datasets: TGIF-QA~\cite{jang2017tgif}, 
MSVD-QA~\cite{xu2017video-msvd}, and MSRVTT-QA~\cite{xu2017video-msvd}, ensuring consistency and comparability with recent multimodal instruction-tuned models. 
For all reported tables, we compute the percentage performance relative to the baseline for each benchmark
and then average these normalized values across all tasks to obtain the Average (Avg.)\%, the last column, thus forming a unified metric for comparison across rows.
\subsection{Results}
\paragraph{Inference-Only Results on LLaVA-1.5-7B.}
Table~\ref{tab:final_inference_table} compares \modelname{} against recent inference-only compression/pruning methods on LLaVA-1.5-7B across five standard benchmarks (POPE, SQA, TextVQA, MME, and GQA). The baseline uses 576 visual tokens, while compressed settings use 192, 128, and 64 tokens (67\%, 78\%, and 89\% token reduction). Across all budgets as shown in table~\ref{tab:trained_method_comparisons}, \modelname{} achieves the strongest accuracy--efficiency trade-off, retaining 99.0\%, 98.1\%, and 95.4\% of baseline average accuracy at 192, 128, and 64 tokens, respectively, outperforming prior methods including FitPrune~\cite{ye2025fit}, SparseVLM~\cite{zhang2024sparsevlm}, VisionZip~\cite{yang2024visionzip}, and PyramidDrop~\cite{xing2024pyramiddrop}. At the matched 192-token setting, \modelname{} reaches 99.0\% relative average accuracy with 1.11$\times$ speedup, improving over VisionZip (97.7\%, 1.14$\times$) and PyramidDrop (96.4\%, 0.97$\times$); under aggressive compression (64 tokens), it still maintains 95.4\% with the highest speedup (1.16$\times$). Notably, at moderate compression, \modelname{} can even surpass baseline performance on POPE and SQA, suggesting improved robustness under token reduction. Overall, these results indicate that the proposed dual-stage reduction strategy preserves the most informative visual/textual tokens and sustains strong cross-modal reasoning as token budgets shrink.
For fairness, the reported results for PyramidDrop are taken directly from their paper, as the authors did not disclose the configuration required to match the average token budgets of 192, 128, and 64 tokens. 
Additionally, since PyramidDrop does not report results on the POPE benchmark, the average performance (Avg.) for this method is computed by averaging across the remaining benchmarks to ensure consistent comparison.

\paragraph{Inference-Only Results on Qwen-2.5-VL-7B.}
Table~\ref{tab:rebuttal_pope_vqa_tokenbudgets} extends our inference-only evaluation to Qwen-2.5-VL-7B, a more recent VLM architecture, and shows that \modelname{} generalizes effectively beyond LLaVA-style backbones. Starting from the dynamic-token baseline, both VisionZip and DUET-VLM (C) provide speedups as token budgets decrease (about 1.3$\times$, 1.4$\times$, and 1.5$\times$ at 640, 320, and 160 average tokens), but DUET-VLM (C) consistently preserves higher task performance. In terms of relative average accuracy, DUET-VLM (C) retains 99.9\%, 99.8\%, and 98.4\% across the three budgets, compared with 99.7\%, 99.1\%, and 96.9\% for VisionZip, with the largest margin under the most aggressive compression. DUET-VLM (C) also remains stronger on challenging metrics such as TextVQA and GQA at lower token budgets, indicating better preservation of informative visual tokens under heavy compression. Overall, these results confirm that DUET is not only effective on earlier architectures, but also robust on recent models such as Qwen-2.5-VL-7B.

\paragraph{Training Results on LLaVA-1.5-7B.}
Table~\ref{tab:train_ablation} presents the comparison of trained methods under different token budgets. Our proposed model variants are compared against VisionZip and PyramidDrop using average token counts of 192, 128, and 64, corresponding to 67\%, 78\%, and 89\% token reduction, respectively. Across all compression levels, \modelname{} consistently achieves superior or comparable performance. At 192 tokens, \modelname{} \crm attains an average accuracy of 99.7\%, closely matching the full-token LLaVA baseline, and even surpasses it on benchmarks such as POPE~\cite{li-etal-2023-evaluating-pope} and GQA~\cite{hudson2019gqa}. As the token budget decreases, our method retains performance more effectively than VisionZip, highlighting the benefit of dual-stage compression. All three variants perform competitively, with \cs slightly outperforming others at lower token budgets.

Table~\ref{tab:trained_method_comparisons} further summarizes the training-time trade-off of DUET-VLM (C). Compared to LLaVA-1.5-7B, DUET-VLM (C) reduces training time by 26\%, 31\%, and 36\% at 192, 128, and 64 tokens, respectively, while maintaining 99.7\% and 99.1\% average accuracy at 192 and 128 tokens. Even under aggressive compression (64 tokens), the model preserves 95.4\% of baseline accuracy, confirming that DUET provides a favorable training efficiency--accuracy balance.

\begin{figure}[!ht]
    \centering
    \includegraphics[width=0.9\columnwidth]{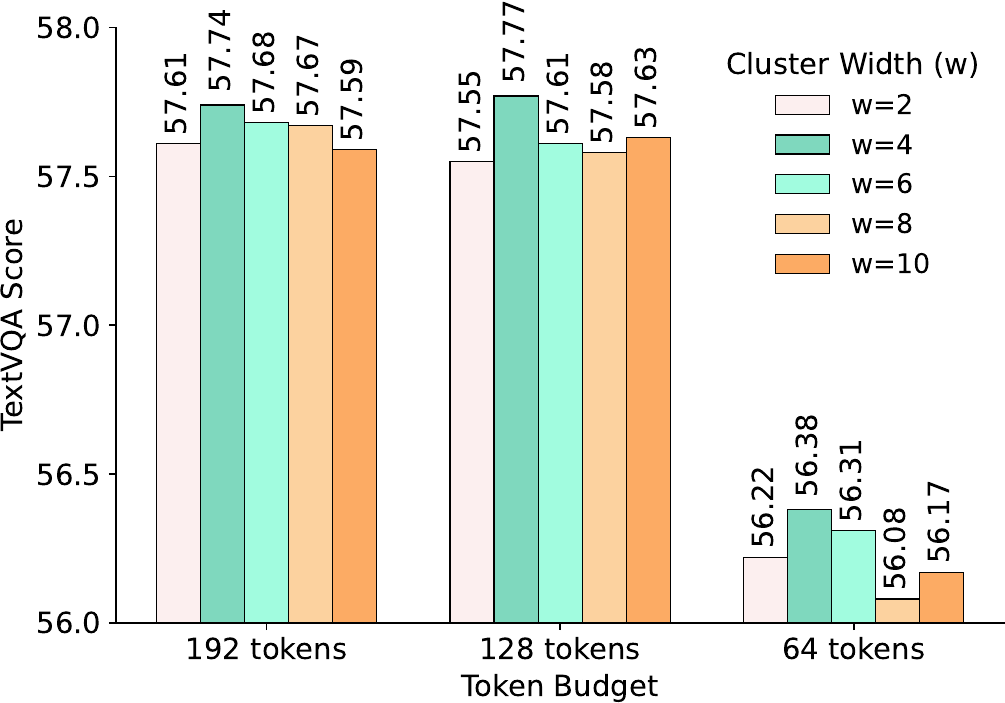}
    \caption{Ablation on varying cluster width of \modelname~\crm on VQA\textsuperscript{T} benchmark on LLaVA-1.5-7B for different token budgets.}
    \label{fig:ablation_cluster_width}
\end{figure}


\paragraph{Inference-Only Results on Video-LLaVA-7B.} Table~\ref{tab:video-llava} presents results on the Video-LLaVA-7B model evaluated across TGIF~\cite{jang2017tgif}, MSVD~\cite{xu2017video-msvd}, and MSRVTT~\cite{xu2017video-msvd} benchmarks 
under two different token budgets.
The full model processes 2048 vision tokens as the upper bound, 
while our \modelname~\crm and other baselines are evaluated at 960 (53.1\% reduction) and 136 (93.4\% reduction) tokens.
At 960 tokens, \modelname~\crm achieves an average accuracy of 100.8\%, even slightly surpassing the baseline, 
demonstrating that our compression retains or enhances task-relevant representations. 
At extreme compression (136 tokens), \modelname~\crm still maintains 97.6\% of the full-model performance, 
substantially outperforming FastV~\cite{chen2024fastv}, SparseVLM~\cite{zhang2024sparsevlm}, and VisionZip by large margins (over 15\% on average). 
These results highlight that video sequences contain significant temporal redundancy and that 
our proposed dual-compression strategy
preserves essential cross-frame information while removing redundant content. 
Overall, \modelname~achieves superior efficiency without compromising multimodal reasoning or video understanding accuracy.



\begin{table}[!ht]
\centering
\tabcolsep=0.08cm
\caption{Comparison showing benefts of our local cluster aggregation on LLaVA-1.5-7B across 6 benchmarks. We report numbers for different methods (M): LLaVA-1.5-7B~(Base), Vanilla (V), VisionZip (VZ), VisionZip with our proposed clustering (VZ (C)), and DUET-VLM (Vanilla) (DV (V)) variants defined in~\cref{subsec:setup}}
\vspace{-2mm}
\label{tab:ablation_local_clustering_llava}
\begin{tabular}{l cccccc c}
\toprule
\multicolumn{1}{l|}{M}      & POPE & SQA\textsuperscript{I}  & VQA\textsuperscript{T} & MME  & GQA  & \multicolumn{1}{c|}{Seed\textsuperscript{I}} & Avg             \\ \midrule
\rowcolor{gray!20}
\multicolumn{8}{c}{Total 576 Tokens (100.0\%)}                                                                  \\
\multicolumn{1}{l|}{Base}      & 85.9 & 69.5 & 58.2    & 1862 & 61.9 & \multicolumn{1}{c|}{66.1} & 100.0\%         \\ \midrule
\rowcolor{gray!20}
\multicolumn{8}{c}{Avg 192 Tokens (~\textcolor{red}{\textdownarrow~66.7\%})}                                                     \\
\multicolumn{1}{l|}{VZ}     & 85.3 & 68.9 & 57.3    & 1783 & 59.3 & \multicolumn{1}{c|}{59.9} & 96.5\%          \\
\multicolumn{1}{l|}{VZ~\crm} & 85.8 & 68.6 & 57.6    & 1743 & 59.0 & \multicolumn{1}{c|}{63.4} & \myuline{97.1\%}    \\ \midrule
\multicolumn{1}{l|}{DV (V)}    & 85.6 & 68.2 & 55.0    & 1802 & 58.8 & \multicolumn{1}{c|}{64.6} & 97.0\%          \\
\multicolumn{1}{l|}{\crm}    & 86.5 & 69.6 & 57.7    & 1817 & 60.3 & \multicolumn{1}{c|}{65.1} & \textbf{98.9\%} \\ \midrule
\rowcolor{gray!20}
\multicolumn{8}{c}{Avg 128 Tokens (~\textcolor{red}{\textdownarrow~77.8\%})}                                                     \\
\multicolumn{1}{l|}{VZ}     & 83.2 & 68.9 & 56.8    & 1762 & 57.6 & \multicolumn{1}{c|}{58.0} & 94.8\%          \\
\multicolumn{1}{l|}{VZ~\crm} & 83.4 & 68.5 & 57.1    & 1748 & 57.6 & \multicolumn{1}{c|}{61.7} & \myuline{95.7\%}    \\ \midrule
\multicolumn{1}{l|}{DV (V)}    & 83.9 & 69.5 & 51.8    & 1679 & 58.1 & \multicolumn{1}{c|}{60.0} & 93.6\%          \\
\multicolumn{1}{l|}{\crm}    & 85.9 & 70.2 & 57.8    & 1767 & 59.0 & \multicolumn{1}{c|}{63.5} & \textbf{97.8\%} \\ \midrule
\rowcolor{gray!20}
\multicolumn{8}{c}{Avg 64 Tokens (~\textcolor{red}{\textdownarrow~88.9\%})}                                                      \\
\multicolumn{1}{l|}{VZ}     & 77.0 & 69.0 & 55.5    & 1690 & 55.1 & \multicolumn{1}{c|}{55.2} & 91.3\%          \\
\multicolumn{1}{l|}{VZ~\crm} & 76.9 & 69.4 & 55.5    & 1706 & 55.4 & \multicolumn{1}{c|}{57.6} & \myuline{92.2\%}    \\ \midrule
\multicolumn{1}{l|}{DV (V)}    & 77.8 & 67.9 & 48.3    & 1565 & 55.5 & \multicolumn{1}{c|}{56.4} & 88.4\%          \\
\multicolumn{1}{l|}{\crm} & 82.5 & 68.3 & 56.4 & 1751 & 56.7 & \multicolumn{1}{c|}{60.7} & \textbf{94.8\%} \\ \bottomrule
\end{tabular}
\vspace{-3mm}
\end{table}

\begin{figure*}[t]
    \centering
    \includegraphics[width=0.95\linewidth]{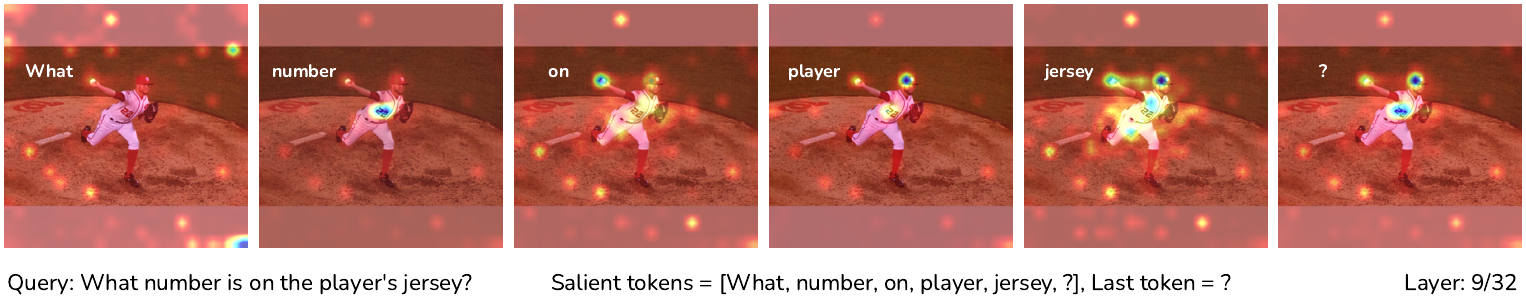}
    \vspace{-2mm}
    \caption{Attention heatmap of salient text tokens attending to visual tokens at the 9\textsuperscript{th} layer of the language backbone in \modelname~\cs.}
    \vspace{-3mm}
    \label{fig:qa_anal}
\end{figure*}

\subsection{Ablations}
\label{subsec:ablations}


\begin{table}[!ht]
\centering
\tabcolsep=0.08cm
\caption{Comparison of \modelname~on LLaVA-1.5-7B under different text token selection schemes for language model-side pruning. We evaluate different Methods (M): LLaVA-1.5-7B~(Base) and our clustering-based variants defined in~\cref{subsec:setup} across different benchmarks. \modelname~maintains strong accuracy even with up to 88.9\% fewer tokens highlighting the effectiveness of our dual-stage visual token compression.}
\vspace{-2mm}
\label{tab:token_sel_llava}
\begin{tabular}{l cccccc c }
\toprule
\multicolumn{1}{l|}{M}         & POPE & SQA\textsuperscript{I}  & VQA\textsuperscript{T} & MME    & GQA  & \multicolumn{1}{c|}{Seed\textsuperscript{I}} & Avg             \\ \midrule
\rowcolor{gray!20}
\multicolumn{8}{c}{Total 576 Tokens (100.0\%)}                                                                             \\ 
\multicolumn{1}{l|}{Base}         & 85.9 & 69.5 & 58.2    & 1862   & 61.9 & \multicolumn{1}{c|}{66.1} & 100.0\%         \\ \midrule
\rowcolor{gray!20}
\multicolumn{8}{c}{Avg 192 Tokens (~\textcolor{red}{\textdownarrow~66.7\%})}                                                                \\
\multicolumn{1}{l|}{\crm}     & 86.5 & 69.6 & 57.7    & 1817 & 60.3 & \multicolumn{1}{c|}{65.1} & \textbf{98.9\%} \\
\multicolumn{1}{l|}{\call} & 86.4 & 68.2 & 57.8    & 1813   & 60.2 & \multicolumn{1}{c|}{65.1} & 98.5\%          \\
\multicolumn{1}{l|}{\cs}   & 86.4 & 68.2 & 57.7    & 1823   & 60.2 & \multicolumn{1}{c|}{65.1} & 98.6\%          \\ \midrule
\rowcolor{gray!20}
\multicolumn{8}{c}{Avg 128 Tokens (~\textcolor{red}{\textdownarrow~77.8\%})}                                                                \\
\multicolumn{1}{l|}{\crm}     & 85.9 & 70.2 & 57.8    & 1767   & 59.0 & \multicolumn{1}{c|}{63.5} & \textbf{97.8\%} \\
\multicolumn{1}{l|}{\call} & 85.9 & 68.4 & 57.8    & 1763   & 59.1 & \multicolumn{1}{c|}{63.5} & 97.3\%          \\
\multicolumn{1}{l|}{\cs}   & 85.9 & 68.4 & 57.7    & 1763   & 59.0 & \multicolumn{1}{c|}{63.6} & 97.3\%          \\ \midrule
\rowcolor{gray!20}
\multicolumn{8}{c}{Avg 64 Tokens (~\textcolor{red}{\textdownarrow~88.9\%})}                                                                 \\
\multicolumn{1}{l|}{\crm}     & 82.5 & 68.3 & 56.4    & 1751   & 56.7 & \multicolumn{1}{c|}{60.7} & 94.8\%          \\
\multicolumn{1}{l|}{\call} & 82.5 & 68.3 & 56.3    & 1753   & 56.6 & \multicolumn{1}{c|}{60.7} & 94.7\%          \\
\multicolumn{1}{l|}{\cs}   & 82.5 & 68.3 & 56.2    & 1757   & 56.6 & \multicolumn{1}{c|}{60.7} & \textbf{94.8\%} \\ \bottomrule
\end{tabular}
\vspace{-3mm}
\end{table}

\begin{table}[!ht]
\centering
\tabcolsep=0.08cm
\caption{Comparison of DUET-VLM on  LLaVA-NeXT-7B under different text token selection schemes for language model-side pruning. We evaluate different methods (M): LLaVA-NeXT-7B~(Base), and our clustering-based variants defined in~\cref{subsec:setup}. \modelname~maintains strong accuracy $>$91\% even at 94.4\% reduction in tokens.}
\vspace{-2mm}
\label{tab:token_sel_llava_next}
\begin{tabular}{@{}lccccccc@{}}
\toprule
\multicolumn{1}{l|}{M}          & POPE & SQA\textsuperscript{I}  & VQA\textsuperscript{T} & MME  & GQA  & \multicolumn{1}{c|}{Seed\textsuperscript{I}} & Avg             \\ \midrule
\rowcolor{gray!20}
\multicolumn{8}{c}{Upper Bound, 2880 Tokens (100.0\%)}                                                                   \\ 
\multicolumn{1}{l|}{Base}         & 86.4 & 70.2 & 61.3    & 1842 & 64.2 & \multicolumn{1}{c|}{70.2} & 100.0\%         \\ \midrule
\rowcolor{gray!20}
\multicolumn{8}{c}{Avg 640 Tokens (~\textcolor{red}{\textdownarrow~77.8\%})}                                                              \\
\multicolumn{1}{l|}{\crm}     & 87.2 & 68.4 & 59.5    & 1839 & 61.8 & \multicolumn{1}{c|}{68.5} & 98.2\%          \\
\multicolumn{1}{l|}{\call} & 87.2 & 68.5 & 59.4    & 1840 & 61.7 & \multicolumn{1}{c|}{68.5} & 98.2\%          \\
\multicolumn{1}{l|}{\cs} & 87.2 & 68.5 & 59.5 & 1839 & 61.6 & \multicolumn{1}{c|}{68.5} & \textbf{98.2\%} \\ \midrule
\rowcolor{gray!20}
\multicolumn{8}{c}{Avg 320 Tokens (~\textcolor{red}{\textdownarrow~88.9\%})}                                                              \\
\multicolumn{1}{l|}{\crm}     & 85.0 & 67.5 & 58.0    & 1771 & 60.1 & \multicolumn{1}{c|}{65.6} & \textbf{95.4\%} \\
\multicolumn{1}{l|}{\call} & 85.1 & 67.0 & 57.8    & 1772 & 60.2 & \multicolumn{1}{c|}{65.7} & 95.3\%          \\
\multicolumn{1}{l|}{\cs}   & 85.0 & 67.1 & 57.9    & 1772 & 60.2 & \multicolumn{1}{c|}{65.7} & 95.3\%          \\ \midrule
\rowcolor{gray!20}
\multicolumn{8}{c}{Avg 160 Tokens (~\textcolor{red}{\textdownarrow~94.4\%})}                                                              \\
\multicolumn{1}{l|}{\crm}     & 80.5 & 67.6 & 57.0    & 1672 & 57.6 & \multicolumn{1}{c|}{61.6} & \textbf{91.8\%} \\
\multicolumn{1}{l|}{\call} & 80.5 & 67.8 & 56.4    & 1670 & 57.6 & \multicolumn{1}{c|}{61.6} & 91.6\%          \\
\multicolumn{1}{l|}{\cs}   & 80.6 & 67.6 & 56.7    & 1670 & 57.6 & \multicolumn{1}{c|}{61.6} & 91.7\%          \\ \bottomrule
\end{tabular}
\vspace{-2mm}
\end{table}

\paragraph{Effect of Local Clustering on LLaVA Models.} Table~\ref{tab:ablation_local_clustering_llava} 
evaluates our local clustering mechanism on LLaVA-1.5-7B.
Across all token budgets, \modelname~\crm consistently surpasses VisionZip, \modelname~(Vanilla), and even VisionZip~\crm--shows clear gains over the original VisionZip demonstrating the general utility of local neighborhood aggregation. 
As shown in Figure~\ref{fig:ablation_cluster_width}, varying cluster width 
(\{4:6\}) yield 
the best TextVQA~\cite{singh2019towards} performance by balancing preservation and redundancy reduction.
as smaller or larger clusters 
cause, either over-fragmentation or over-smoothing. 
Overall, \modelname~\crm achieves up to 97.1\% of baseline accuracy under heavy compression, validating
that localized clustering is key to compact yet semantically rich token representation.

\paragraph{Effect of Text-Guided Token Selection in \modelname.} Tables~\ref{tab:token_sel_llava} and~\ref{tab:token_sel_llava_next} present an ablation for different token selection strategies 
within the language backbone of \modelname~for both LLaVA-1.5-7B and LLaVA-NeXT-7B. 
For both models and all token budgets, \modelname~\cs consistently performs on par with or better than the other variants, particularly at higher compression levels.
For instance, on LLaVA-1.5-7B, \modelname~\cs achieves 98.9\%, 97.8\%, and 94.8\% of baseline accuracy at 192, 128, and 64 tokens, respectively, 
showing strong retention even with 89\% token reduction. 
Similarly, on LLaVA-NeXT-7B, the \cs variant attains 98.2\%, 95.4\%, and 91.7\% accuracy at 640, 320, and 160 tokens, respectively, matching or slightly 
exceeding the \call configuration across all benchmarks. 
The advantage of \cs lies in its `selective focus': salient text tokens provide sharper contextual cues for identifying 
relevant visual regions, avoiding the over-smoothing effect introduced when all text tokens are used equally.
This targeted text guidance enhances token importance estimation, allowing the model to preserve semantically rich patches 
while pruning redundant ones more effectively. 
These results validate that coupling text-guided saliency with visual clustering
is a key factor behind \modelname’s superior efficiency–accuracy trade-off and represents one of the core contributions of this work.

\paragraph{Qualitative Analysis.}
From the attention heatmap in \cref{fig:qa_anal}, we observe that the salient text tokens (including the last text token) guide the token-dropping mechanism in \modelname, narrowing the model’s focus toward query-relevant image regions. 
This selective emphasis helps remove irrelevant background and distracting objects, leading to more accurate visual reasoning.

\section{Conclusion}
\label{sec:conclusion}
In this work, we introduced \modelname, a dual-stage visual compression framework that achieves efficient and accurate multimodal learning under aggressive token reduction.
Our key insight is that redundancy exists both spatially in vision tokens and contextually during multimodal fusion.
By combining redundancy-aware visual token merging in the vision encoder with saliency-based text-guided token pruning in the language backbone, 
\modelname~achieves substantial compression while retaining semantically critical information.
Experiments on image and video benchmarks show that \modelname~maintains near-baseline accuracy
in inference-only settings and even surpasses baseline performance when fine-tuned under the same compression regime across all benchmarks.
These findings demonstrate that training-aware token management--not model scaling--is a promising path toward compact yet high-performing VLMs.
A comprehensive sensitivity analysis of parameters $k_1, k_2, \lambda$, and $l$ is provided in the supplemental.
We do not report exact inference or training times, but our runtimes are comparable to VisionZip~\cite{yang2024visionzip} under the same token budget.

A detailed speed analysis is left for future work, as we aim to develop optimized kernels to further accelerate both inference and training. 
We also leave the training of video models and inference on longer-horizon videos (with more frames than our current setting) to future work, enabled by the minimal accuracy drop observed even under extreme token compression with our approach.
For more precise visual token reduction in vision encoder, future work may explore stronger region-proposal or saliency-based methods beyond the current V2V self-attention filtering.
Finally, we would like to extend this dual-stage vision compression paradigm to additional modalities (\eg, audio and text) and further advance practical, scalable multimodal systems.


{
   \small
   \bibliographystyle{ieeenat_fullname}
   \bibliography{bib/longstrings,main}
}

\clearpage
\appendix

\begin{figure*}[!htbp]
    \centering
    \includegraphics[width=\linewidth]{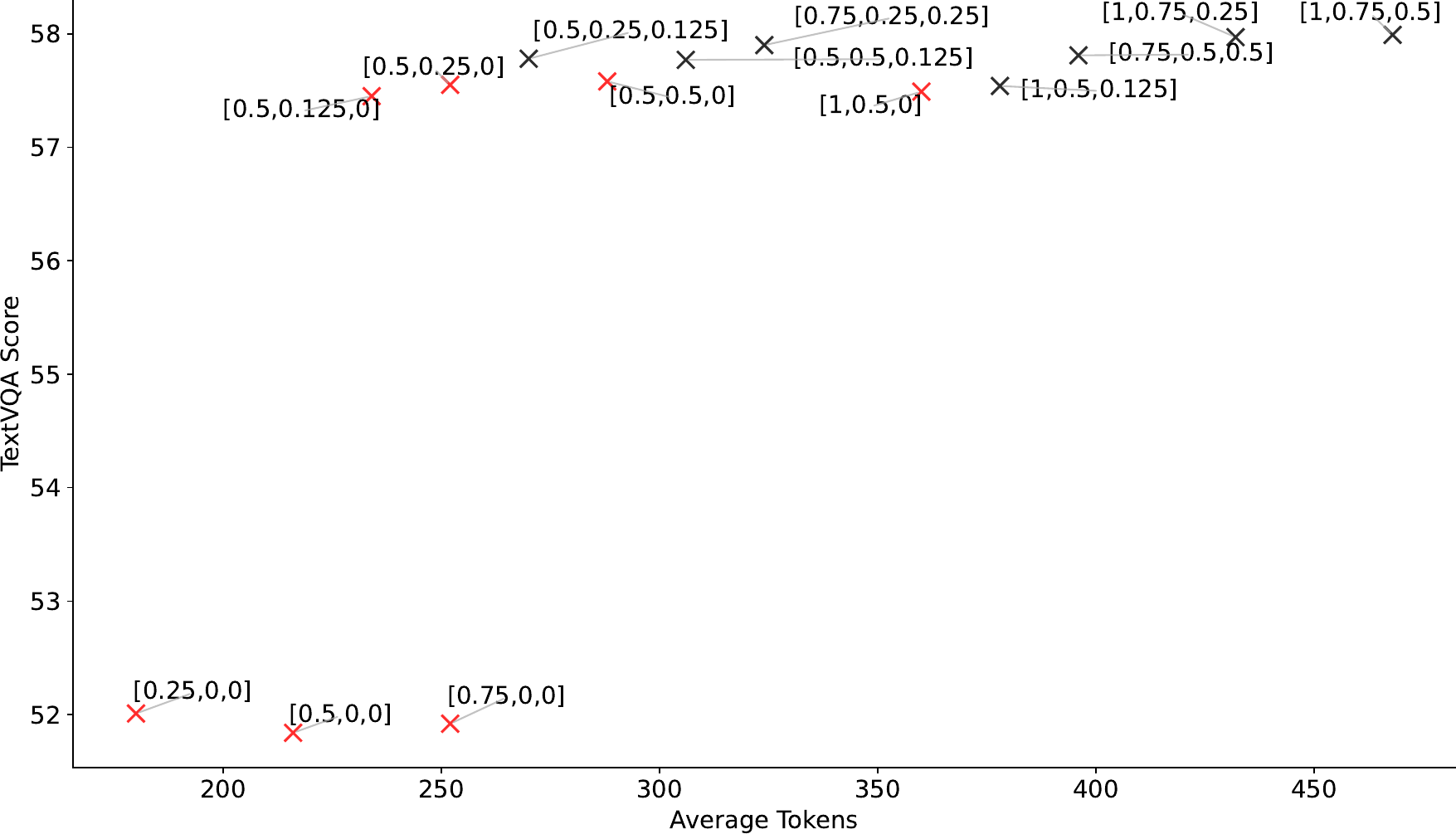}
    \vspace{2mm}
    \caption{Performance of various dropping configurations in the language backbone on TextVQA using all text tokens. Red crosses indicate configurations in which all visual tokens are removed at some layer prior to the final layer.}
    \label{fig:rank_drop_cfg}
\end{figure*}

\section{Sensitivity Analysis}

\begin{figure*}[!htbp]
    \centering
    \includegraphics[width=0.9\linewidth]{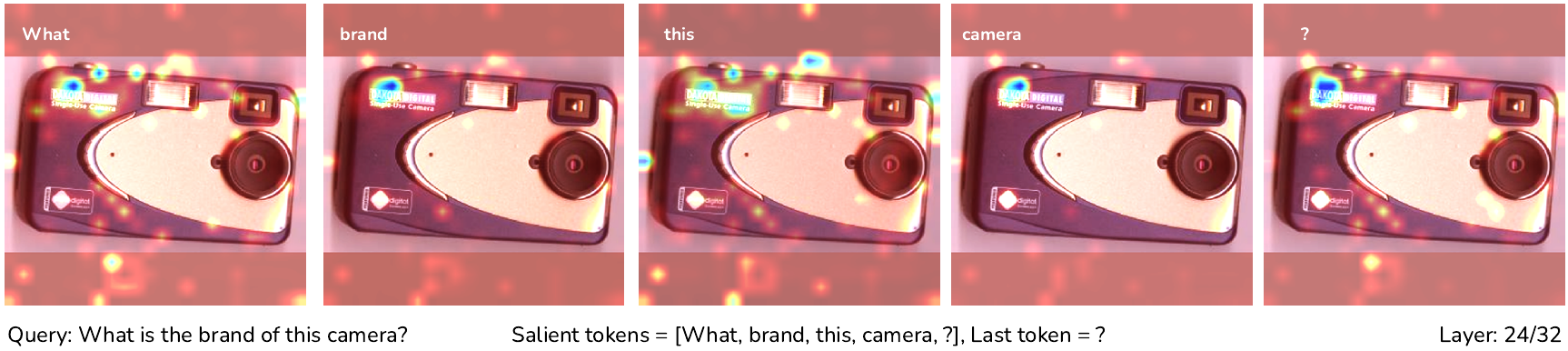}
    \caption{Attention heatmap of salient text tokens attending to visual tokens at the 24\textsuperscript{th} layer of the language backbone in \modelname~\cs.}
    \label{fig:qa_anal_suppl}
\end{figure*}

\paragraph{Effect on varying parameters $\lambda$ and $l$.} 
To study the behavior of our token-dropping mechanism, we analyze the effect of the compression ratio $\lambda$ and the stage index $l$ in~\cref{fig:rank_drop_cfg}. 
Here, $\lambda \in [0,1]$ denotes the fraction of vision tokens dropped at each stage, and $l \in \{1,2,3\}$ indexes the segmentation stages of the language backbone. 
In all experiments, we fix the three segmentation points at $[8,16,24]$ layers, corresponding to the three stages, and vary $\lambda$ to evaluate the sensitivity of our method.
We observe that pruning tokens entirely in the final layers has only a minimal effect on performance. 
The red crosses denote configurations in which all vision tokens are pruned starting in one of three stages.
When this 100\% pruning occurs in the middle stage, i.e., where $\lambda = 0$ for all layers after the 16\textsuperscript{th} layer performance degrades because the visual information has not yet been fully extracted into the hidden states. 
However, in the later layers (last stage), this information has already been extracted, making the vision tokens redundant. 
Consequently, for all configurations where 100\% pruning occurs after the 24\textsuperscript{th} layer, the performance remains competitive. A simple vizualization of this can be seen in~\cref{fig:qa_anal_suppl} where at the 24\textsuperscript{th} layer all the salient tokens as well as the last token focus on the required region, which implies that the required knowledge of the image has been transferred to the hidden states and the vision tokens are now redundant.
We follow the 8-layer segmentation used in PyramidDrop~\cite{xing2024pyramiddrop} for this analysis. A more fine-grained, layer-by-layer investigation is left for future work.

\begin{figure*}[!htbp]
    \centering
    \includegraphics[width=\linewidth]{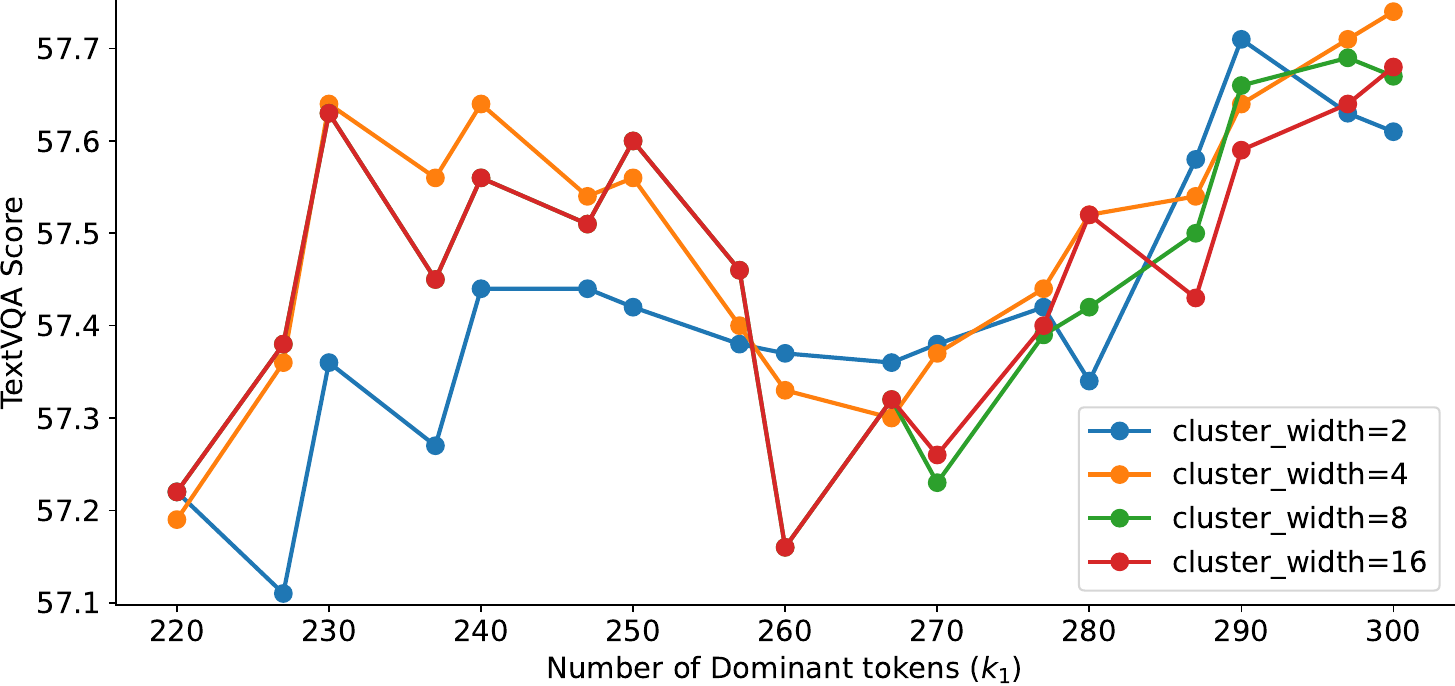}
    \vspace{2mm}
    \caption{Plot of TextVQA performance across varying $k_1$ for different cluster widths, under the constraint $k_1+k_2=$ 307 needed to meet a token budget of 192 using \modelname~\crm on LLaVA-1.5-7B.}
    \label{fig:cluster_cfg_192}
\end{figure*}

\begin{figure*}[!htbp]
    \centering
    \includegraphics[width=\linewidth]{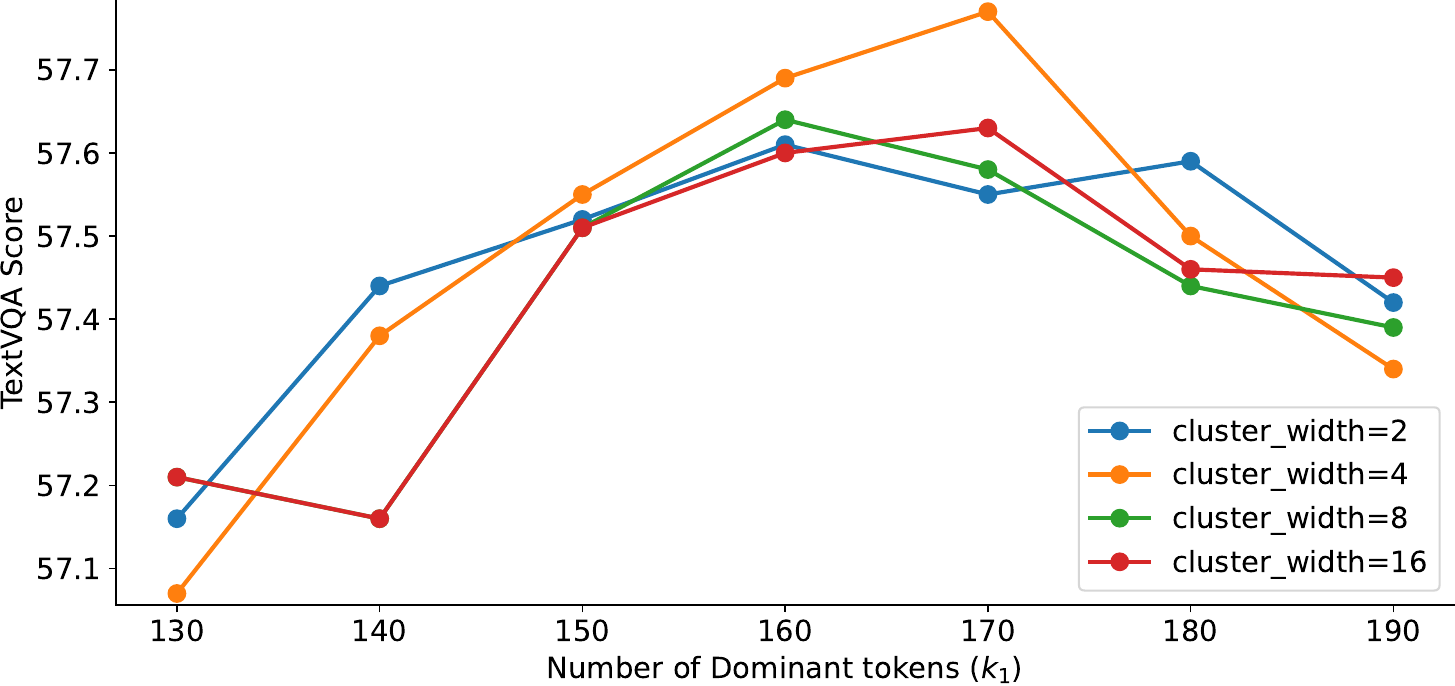}
    \caption{Plot of TextVQA performance across varying $k_1$ for different cluster widths, under the constraint $k_1+k_2=$ 205 needed to meet a token budget of 128 using \modelname~\crm on LLaVA-1.5-7B.}
    \label{fig:cluster_cfg_128}
\end{figure*}


\paragraph{Effect on varying parameters $k_1$ and $k_2$.} 
We analyze the effect of varying the number of dominant ($k_1$) and contextual ($k_2$) tokens across different cluster widths for a couple of token budgets as shown in~\cref{fig:cluster_cfg_192} and~\cref{fig:cluster_cfg_128} respectively. Since $k_1 + k_2$ is fixed by the token budget, we visualize only the variation in $k_1$; the corresponding trend for $k_2$ would be a symmetric reflection.
Interestingly, the trends in Fig.~6 and Fig.~7 are not identical, and this difference directly informed our configuration choices. In~\cref{fig:cluster_cfg_192} (192-token budget), we observe a clear upward trend as the number of dominant tokens $k_1$ increases toward the maximum feasible value ($\approx$300), performance improves consistently across all cluster widths. This indicates that, at a relatively relaxed token budget, the model benefits from allocating a large dominant token pool, as it preserves more high-saliency visual information.
However, in~\cref{fig:cluster_cfg_128} (128-token budget), the trend changes. Here, performance peaks at a mid-range of dominant tokens (around 155–165) and then drops when $k_1$ becomes too large. This suggests that at stricter budgets, dominating the token set with high-saliency visual tokens ($k_1$) reduces contextual diversity and harms reasoning performance.
Together, these results show that:
\begin{itemize}[leftmargin=*]
    \item At higher budgets (192 tokens) $\longrightarrow$ more dominant tokens ($k_1$) are consistently better.

    \item At lower budgets (128 tokens) $\longrightarrow$ an intermediate dominant–contextual ($k_1-k_2$) split yields the optimal balance.

\end{itemize}
These observations guided our chosen configurations, we scale $k_1$ upward for larger budgets, while adopting a balanced dominant-contextual ratio ($k_1:k_2$) for more aggressive compression scenarios.

\section{Additional Experiments}

\begin{table}[!ht]
\centering
\tabcolsep=0.078cm
\caption{Comparison showing benefts of our local cluster aggregation on LLaVA-NeXT-7B across 6 benchmarks. We report numbers for different methods (M): LLaVA-NeXT-7B~(Base), Vanilla (V), VisionZip (VZ), VisionZip with our proposed clustering (VZ (C)), DUET-VLM (Vanilla) (DV (V)) and our proposed method, DUET-VLM (C) represented as (C)}
\label{tab:ablation_local_clustering_llava_next}
\begin{tabular}{l cccccc c}
\toprule
\multicolumn{1}{l|}{M}      & POPE & SQA\textsuperscript{I}  & VQA\textsuperscript{T} & MME  & GQA  & \multicolumn{1}{c|}{Seed\textsuperscript{I}} & Avg             \\ \midrule
\rowcolor{gray!20}
\multicolumn{8}{c}{Upper Bound, 2880 Tokens (100\%)}                                                            \\
\multicolumn{1}{l|}{V}      & 86.4 & 70.2 & 61.3    & 1842 & 64.2 & \multicolumn{1}{c|}{70.2} & 100.0\%         \\ \midrule
\rowcolor{gray!20}
\multicolumn{8}{c}{Avg 640 Tokens (~\textcolor{red}{\textdownarrow~77.8\%})}                                                     \\
\multicolumn{1}{l|}{VZ}     & 86.3 & 68.1 & 60.2    & 1787 & 61.3 & \multicolumn{1}{c|}{66.7} & 97.1\%          \\
\multicolumn{1}{l|}{VZ~\crm} & 86.4 & 67.8 & 60.1    & 1786 & 61.3 & \multicolumn{1}{c|}{66.9} & {\ul 97.1\%}    \\ \midrule
\multicolumn{1}{l|}{DV (V)}    & 85.0 & 67.8 & 50.3    & 1750 & 60.7 & \multicolumn{1}{c|}{65.8} & 93.4\%          \\
\multicolumn{1}{l|}{\crm} & 87.2 & 68.4 & 59.5 & 1839.0 & 61.8 & \multicolumn{1}{c|}{68.5} & \textbf{98.2\%} \\ \midrule
\rowcolor{gray!20}
\multicolumn{8}{c}{Avg 320 Tokens (~\textcolor{red}{\textdownarrow~88.9\%})}                                                     \\
\multicolumn{1}{l|}{VZ}     & 82.1 & 67.3 & 58.9    & 1702 & 59.3 & \multicolumn{1}{c|}{63.4} & {\ul 93.7\%}    \\
\multicolumn{1}{l|}{VZ~\crm} & 82.1 & 67.4 & 59.1    & 1670 & 59.1 & \multicolumn{1}{c|}{63.5} & 93.4\%          \\ \midrule
\multicolumn{1}{l|}{DV (V)}    & 80.6 & 68.2 & 47.0    & 1612 & 59.1 & \multicolumn{1}{c|}{63.2} & 89.5\%          \\
\multicolumn{1}{l|}{\crm}    & 85.0 & 67.5 & 58.0    & 1771 & 60.1 & \multicolumn{1}{c|}{65.6} & \textbf{95.4\%} \\ \midrule
\rowcolor{gray!20}
\multicolumn{8}{c}{Avg 160 Tokens (~\textcolor{red}{\textdownarrow~94.4\%})}                                                     \\
\multicolumn{1}{l|}{VZ}     & 74.8 & 68.3 & 56.2    & 1630 & 55.5 & \multicolumn{1}{c|}{58.3} & {\ul 88.9\%}    \\
\multicolumn{1}{l|}{VZ~\crm} & 75.1 & 67.4 & 56.2    & 1650 & 54.9 & \multicolumn{1}{c|}{58.3} & 88.8\%          \\ \midrule
\multicolumn{1}{l|}{DV (V)}    & 75.2 & 68.4 & 44.7    & 1576 & 57.8 & \multicolumn{1}{c|}{61.4} & 86.7\%          \\
\multicolumn{1}{l|}{\crm}    & 80.5 & 67.6 & 57.0    & 1672 & 57.6 & \multicolumn{1}{c|}{61.6} & \textbf{91.8\%} \\ \bottomrule
\end{tabular}
\vspace{-3mm}
\end{table}

\begin{table}[ht]
\centering
\tabcolsep=0.14cm
\caption{Comparison of various token-selection schemes applied to PyramidDrop (PDrop) on top of LLaVA-1.5-7B (Base)}
\label{tab:token_sel_pdrop}

\begin{tabular}{l ccccc }
\toprule
\multicolumn{1}{l|}{Method} &
  VQA\textsuperscript{T} &
  POPE &
  SQA\textsuperscript{I} &
  \multicolumn{1}{l|}{MME} &
  Avg \\ \midrule
\rowcolor{gray!20}
\multicolumn{6}{c}{Total 576 Tokens (100\%)} \\ \midrule
\multicolumn{1}{l|}{Base} &
  \begin{tabular}[c]{@{}l@{}}58.2 \end{tabular} &
  \begin{tabular}[c]{@{}l@{}}85.9\end{tabular} &
  \begin{tabular}[c]{@{}l@{}}69.5\end{tabular} &
  \multicolumn{1}{l|}{\begin{tabular}[c]{@{}l@{}}1862\end{tabular}} &
  100.0\% \\ \midrule
\rowcolor{gray!20}
\multicolumn{6}{c}{Avg 270 Tokens (~\textcolor{red}{\textdownarrow~53.1\%})} \\ \midrule
\multicolumn{1}{l|}{PDrop} &
  \begin{tabular}[c]{@{}l@{}}57.5\end{tabular} &
  \begin{tabular}[c]{@{}l@{}}84.8\end{tabular} &
  \begin{tabular}[c]{@{}l@{}}69.4\end{tabular} &
  \multicolumn{1}{l|}{\begin{tabular}[c]{@{}l@{}}1854.0\end{tabular}} &
  99.2\% \\
\multicolumn{1}{l|}{PDrop (S)} &
  \begin{tabular}[c]{@{}l@{}}57.6\end{tabular} &
  \begin{tabular}[c]{@{}l@{}}85.2\end{tabular} &
  \begin{tabular}[c]{@{}l@{}}69.2\end{tabular} &
  \multicolumn{1}{l|}{\begin{tabular}[c]{@{}l@{}}1862\end{tabular}} &
  99.4\% \\
\multicolumn{1}{l|}{PDrop (all)} &
  \begin{tabular}[c]{@{}l@{}}57.8\end{tabular} &
  \begin{tabular}[c]{@{}l@{}}85.2\end{tabular} &
  \begin{tabular}[c]{@{}l@{}}69.2\end{tabular} &
  \multicolumn{1}{l|}{\begin{tabular}[c]{@{}l@{}}1862.0\end{tabular}} &
  \textbf{99.5\%} \\ \bottomrule
\end{tabular}%
\vspace{-3mm}
\end{table}
\paragraph{Effect of local clustering on LLaVA-NeXT-7B.}
Across all token budgets, we consistently observe that local clustering strengthens the performance of both VisionZip (VZ) and DUET-VLM (DV), demonstrating its robustness as a plug-in compression strategy. At the 640-token budget ($~$78\% reduction), clustering provides a modest yet stable improvement, indicating that early-stage redundancy is already substantial and can be effectively leveraged. As the budget becomes more restrictive (320 and 160 tokens), the benefit of clustering becomes even more pronounced: VisionZip~\crm and \modelname~\crm retain significantly higher accuracy compared to their non-clustered counterparts, with \modelname~\crm showing the strongest resilience in the extremely low-token regime. This trend highlights that preservation of local structure is especially crucial when aggressively compressing vision features allowing the model to retain the most semantically coherent groups of tokens rather than isolated high-scoring ones. Overall, the results confirm that local clustering provides a consistent, compression-aware advantage that amplifies the effectiveness of both VisionZip and DUET-VLM, particularly under severe token constraints.

\paragraph{Effect of Text Token Selection in PyramidDrop.} \Cref{tab:token_sel_pdrop} compares different strategies for selecting text tokens used to guide visual token ranking and pruning within PyramidDrop. 
The baseline approach of PyramidDrop relies on the \emph{last text token} for ranking, while we propose two alternatives: 
\emph{salient text tokens} (those with the highest attention weights) and \emph{all query text tokens}. 
Both variants yield consistent improvement in accuracy across different benchmarks, demonstrating that leveraging broader and more relevant textual context 
enhances the model’s ability to estimate token importance and retain semantically relevant visual features. 
Using all query tokens achieves the highest average accuracy (99.5\%), marginally outperforming the salient-token variant (99.4\%), 
suggesting that distributing attention across the full text sequence can better capture fine-grained multimodal relationships. 
Although the performance gap between the two strategies is minimal, these findings confirm that integrating textual guidance 
into the ranking mechanism improves information retention without incurring additional computational cost. 

\section{Hyperparameters}
\vspace{-3mm}
\begin{table}[ht]
\caption{Vision token configuration used in the local clustering method for LLaVA-1.5-7B to achieve different target tokens}
\vspace{-2mm}
\label{tab:llava-cfg}
\centering
\resizebox{0.7\columnwidth}{!}{%
\begin{tabular}{@{}c|cc@{}}
\toprule
Target Tokens & Dominant & Contextual \\ \midrule
192           & 300      & 7          \\
128           & 170      & 35         \\
64            & 72       & 30         \\ \bottomrule
\end{tabular}%
}
\end{table}
\begin{table}[ht]
\caption{Vision token configuration used in the local clustering for LLaVA-NeXT-7B to achieve different target tokens}
\vspace{-2mm}
\label{tab:llava-next-cfg}
\centering
\resizebox{0.7\columnwidth}{!}{%
\begin{tabular}{@{}c|cc@{}}
\toprule
Target Tokens & Dominant & Contextual \\ \midrule
640           & 850      & 175          \\
320           & 360      & 150         \\
160           & 225      & 30         \\ \bottomrule
\end{tabular}%
}
\end{table}
\begin{table}[ht]
\caption{Vision token configuration used in the local clustering for Video-LLaVA-7B to achieve different target tokens}
\label{tab:video-llava-cfg}
\centering
\resizebox{0.7\columnwidth}{!}{%
\begin{tabular}{@{}c|cc@{}}
\toprule
Target Tokens & Dominant & Contextual \\ \midrule
960           & 1280      & 256          \\
136           & 160      & 56         \\
\bottomrule
\end{tabular}%
}
\end{table}
\begin{table}[t]
\centering
\begin{tabular}{c|c c}
\toprule
Target Tokens & Dominant & Contextual \\
\midrule
640 & 870 & 153 \\
320 & 517 & 91 \\
160 & 207 & 36 \\
\bottomrule
\end{tabular}
\caption{Vision token configuration used in local clustering for Qwen2.5-VL-7B to achieve different target token budgets. The LLM-side rank-and-drop  uses $\texttt{layer\_list}=[14,21]$ and $\texttt{image\_token\_ratio\_list}=[0.5,0.25]$.}
\label{tab:qwen_local_clustering_config}
\end{table}
\paragraph{Token Numbers for Local Clustering.}
\noindent\cref{tab:llava-cfg} reports the counts of dominant and contextual tokens selected for LLaVA-1.5-7B to meet three target token budgets after pruning on the language backbone side. Similarly, \cref{tab:llava-next-cfg} presents these counts for LLaVA-NeXT-7B across three configurations. 
\cref{tab:video-llava-cfg} summarizes the dominant and contextual token counts for Video-LLaVA-7B, which processes eight video frames to capture temporal context. These configurations are guided by the trends in~\cref{fig:cluster_cfg_192} and~\cref{fig:cluster_cfg_128}, where higher token budgets favor the allocation of more dominant tokens, while tighter budgets achieve the best performance with a balanced dominant–contextual split.


\end{document}